\documentclass[pdflatex,sn-mathphys,iicol]{sn-jnl}%

\jyear{2021}%

\theoremstyle{thmstyleone}%

\theoremstyle{thmstyletwo}%

\theoremstyle{thmstylethree}%

\usepackage{amsfonts}
\usepackage{amsmath}
\usepackage{amssymb}
\usepackage{apxproof}
\usepackage{booktabs}
\usepackage{caption}
\usepackage{graphicx}
\usepackage{microtype}
\usepackage{nicefrac}
\usepackage{subcaption}
\usepackage{url}
\usepackage{xcolor}
\usepackage{xspace}
\usepackage{wrapfig}
\usepackage{fontawesome}
\usepackage[capitalize]{cleveref}

\makeatletter
\DeclareRobustCommand\onedot{\futurelet\@let@token\@onedot}
\def\@onedot{\ifx\@let@token.\else.\null\fi\xspace}

\def\eg{\emph{e.g}\onedot}

\makeatother

\DeclareMathOperator*{\argmin}{arg\,min}

\newcommand{\tinyspace}{\kern 0.025em}
\newcommand{\stemp}{\faChild\@\xspace}
\newcommand{\skey}{\faKey\@\xspace}
\newcommand{\sview}{\faCamera\@\xspace}
\newcommand{\smask}{\faScissors\@\xspace}
\newcommand{\svid}{\faVideoCamera\@\xspace}
\newcommand{\sflow}{\faForward\@\xspace}

\newcommand{\supcmr}{\sview\tinyspace\skey\tinyspace\smask}
\newcommand{\supucmr}{\stemp\tinyspace\smask}
\newcommand{\supumr}{\smask}
\newcommand{\supvmr}{\stemp\tinyspace\sview\tinyspace\skey\tinyspace\smask}
\newcommand{\supours}{\smask\tinyspace\sflow\tinyspace\svid}

\normalbaroutside

\begin{document}

\title[Article Title]{DOVE\@: Learning Deformable 3D Objects by Watching Videos}

\author*[1]{\fnm{Shangzhe} \sur{Wu}}\email{szwu@robots.ox.ac.uk}
\equalcont{These authors contributed equally to this work.}

\author*[1]{\fnm{Tomas} \sur{Jakab}}\email{tomj@robots.ox.ac.uk}
\equalcont{These authors contributed equally to this work.}

\author[1]{\fnm{Christian} \sur{Rupprecht}}\email{chrisr@robots.ox.ac.uk}

\author[1]{\fnm{Andrea} \sur{Vedaldi}}\email{vedaldi@robots.ox.ac.uk}

\affil[1]{\orgdiv{Visual Geometry Group}, \orgname{University of Oxford}}

\abstract{
Learning deformable 3D objects from 2D images is often an ill-posed problem.
Existing methods rely on explicit supervision to establish multi-view correspondences, such as template shape models and keypoint annotations, which restricts their applicability on objects ``in the wild''.
A more natural way of establishing correspondences is by watching videos of objects moving around.
In this paper, we present DOVE, a method that learns textured 3D models of deformable object categories from monocular videos available online, without keypoint, viewpoint or template shape supervision.
By resolving symmetry-induced pose ambiguities and leveraging temporal correspondences in videos, the model automatically learns to factor out 3D shape, articulated pose and texture from each individual RGB frame, and is ready for single-image inference at test time.
In the experiments, we show that existing methods fail to learn sensible 3D shapes without additional keypoint or template supervision, whereas our method produces temporally consistent 3D models, which can be animated and rendered from arbitrary viewpoints.
Project page: {\url{https://dove3d.github.io/}}.
}


\maketitle

\begin{figure*}[t]
    \centering
    \includegraphics[trim={0 0 30px 0}, clip, width=\linewidth]{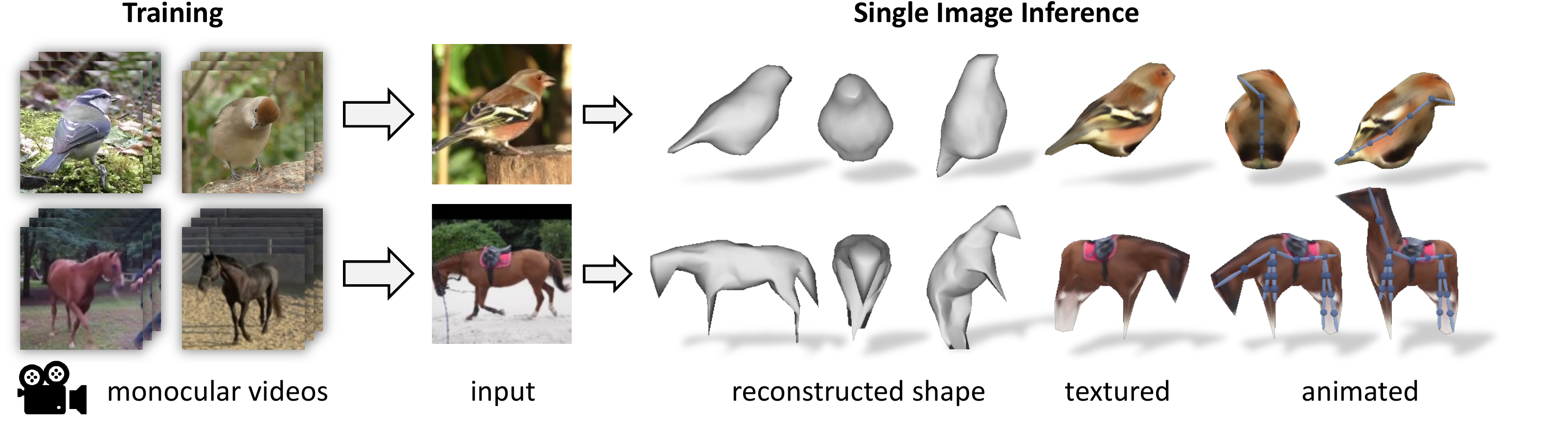}
    \vspace{0.03in}
    \caption{\textbf{DOVE - Deformable Objects from VidEos.} Given a collection of video clips of an object category as training data, we learn a model that is able to predict a textured, articulated 3D mesh of the object from a single input image.}
    \label{fig:teaser}
\end{figure*}
\section{Introduction}\label{s:intro}

In applications, we often need to obtain accurate 3D models of deformable objects from just one or a few pictures of them.
This is the case in traditional applications such as robotics, but also, increasingly, in consumer applications, such as content creation for virtual and augmented reality --- using everyday pictures and videos taken with a cellphone.

3D reconstruction from a single image, or even a small number of views, is generally very ambiguous and only solvable by leveraging powerful statistical priors of the 3D world.
Learning such priors is however very challenging.
One approach is to use training data specifically collected for this purpose, for example by using 3D scanners and domes~\cite{choy20163d, Girdhar16b, Wu163dgan, wang2018pixel2mesh, groueix2018, kato2018renderer, occnet, Park_2019_CVPR, pifuSHNMKL19, meshrcnn} or shape models~\cite{loper15smpl:, hmrKanazawa17, Zuffi:CVPR:2017, RingNet:CVPR:2019, Zuffi19Safari}.
This is expensive and can be justified only for a few categories such as human bodies and faces that are of particular significance in applications.
However, scanning is not a viable approach to cover the huge diversity of objects that exist in the real world.

We thus need to develop methods that can learn 3D deformable objects from as cheap supervision as possible, such as leveraging casually-collected images and videos found on the Internet, or crowdsourced datasets such as CO3D~\cite{reizenstein2021common}.
Ideally, our system should take as input a collection of such casual images and videos and learn a model capable of reconstructing the 3D shape, appearance and deformation of a new object from a single image of it.

While several authors have looked at this problem before~\cite{choy20163d, Girdhar16b, wang2018pixel2mesh, groueix2018, kato2018renderer, meshrcnn}, so far it has always been necessary to make additional simplifying assumptions compared to the ideal unsupervised setting described above.
These assumptions usually come in the form of additional geometric supervision.
The most common one is to require 2D masks for the objects, either obtained manually or via a pre-trained segmentation network such as~\cite{he17mask, kirillov2019pointrend}.
On top of this, there is usually at least one more \emph{additional} form of geometric supervision, such as providing an initial approximate 3D template of the object, 2D keypoint detections, or approximate 3D camera parameters~\cite{kanazawa18learning, li20online, Kokkinos_2021_CVPR, kulkarni19canonical, Zuffi19Safari, goel20shape, kokkinos2021point, DVR}.
There is a small number of works that require no masks or geometric supervision~\cite{wu20unsupervised}, but they come with other limitations such as relying on limited viewpoint range.

Our aim in this paper is to learn 3D deformable objects from complex \emph{casual videos} while only using 2D masks and removing the additional geometric supervision that require expensive manual annotations used in prior works (keypoints, viewpoint, and templates).
In order to compensate for this lack of geometric information, we propose to learn from casual videos rather than still images, unlike most prior works.
While this adds some complexity to the method, using videos has the key advantage to allow one to establish correspondences between different images, for instance by using an off-the-shelf optical flow algorithm.
While this information is weaker than externally-provided information such as keypoints, nevertheless it is very helpful in recovering the objects' viewpoint.
Note, though, that videos are only used for supervision: our goal is still to learn a model that can reconstruct a new object instance from a single image.

In order to use videos effectively, we make a number of technical contributions.
The first one addresses the challenge of estimating the viewpoint of the 3D objects.
Prior works addressed this issue by sampling a large number of possible views~\cite{kulkarni20articulation-aware,goel20shape}, an approach that~\cite{goel20shape} calls a \emph{camera multiplex}.
We find that this is unnecessary.
While viewpoint estimation is ambiguous, we show that the ambiguity  is mostly restricted to a small space of symmetries induced by the 2D projection of the 3D objects onto the image.
The result is that, as the model is learned, only a very small number of alternative viewpoints need to be explored in order to escape form the ambiguity-induced local optima: from, e.g., 40 in~\cite{goel20shape} to just two per iteration, which largely reduces memory and time requirements for training.

Our second contribution is the design of the object model.
We propose a hierarchical shape representation that explicitly disentangles category-dependent prior shape, instance-dependent deformation, as well as time-dependent articulated and rigid pose.
In this way, we automatically factor shape and pose variations at different levels in the video dataset, and leverage instance-specific correspondences within a video and instance-agnostic correspondences across multiple videos.
We also enforce a bilateral symmetry on the predicted canonical shape and texture, similar to previous methods~\cite{kanazawa18learning,goel20shape,li20self-supervised,wu20unsupervised}.
However, differently from these approaches, which assume symmetry at the level of the object instances, here we assume the canonical (pose free) shapes are symmetric, but individual articulations can be asymmetric~\cite{thewlis18modelling,fernandez-labrador20unsupervised}, which is much more realistic.

We also address the challenge of evaluating these reconstruction methods.
Prior works in this area generally lack data with 3D ground truth.
Instead, they resort to indirect evaluation by measuring the quality of the 2D correspondences that are established by the 3D models.
To address this problem, we create a dataset of views of real-life animal models (toy birds).
The data is designed to resemble a subset of the images as found in existing datasets such as CUB~\cite{WahCUB_200_2011};
however, it additionally comes with 3D scans of the objects, which can be used to test the quality of the 3D reconstructions directly.
We use this data to evaluate our and several state-of-the-art algorithms without the need for proxy metrics such as keypoint re-projection error that are insufficient to assess the quality of a 3D reconstruction.
We hope that this data will be useful for future work in this area.

Overall, our method can successfully learn good 3D shape predictors from videos of animals such as birds and horses.
Compared to prior work, our method produces better 3D shape reconstructions, as measured on the new benchmark, when not using additional geometric supervision.

\begin{table*} [t!]
    \setlength{\tabcolsep}{1.5pt}
    \small
	\centering
    \caption{\textbf{Related Work Overview.} Annotations: \stemp template shape, \sview viewpoint, \skey 2D keypoint, \smask object mask, \sflow optical flow, \svid video, $^*$optimizes a single object instance over a single or a few sequences, $^1$shape bases initialized from CMR~\cite{kanazawa18learning}, $^2$outputs texture flow, $^3$obtained from DensePose~\cite{neverova20continuous}, $^4$obtained from keypoints using SfM, $^\dagger$UMR~\cite{li20self-supervised} relies on part segmentations from SCOPS~\cite{hung19scops:}.
    }
	\begin{tabular}{lccccccccccc}
		\toprule
		& \multicolumn{6}{c}{Supervision} & \multicolumn{5}{c}{Output} \\
		\cmidrule(lr){2-7} \cmidrule(lr){8-12}
		Method                              &    \stemp        &  \sview            & \skey  &  \smask         & \sflow        & \svid       & 3D            & 2.5D         & Motion       & Viewpoint         & Texture           \\
        \midrule
        VMR$^*$~\cite{li20online}           & (\checkmark)$^1$ &  &  & \checkmark  &             & \checkmark  & \checkmark    &              & \checkmark   & \checkmark   & (\checkmark)$^2$ \\
        LASR$^*$~\cite{yang21lasr:}                    &                   &                  &            &  \checkmark & \checkmark  & \checkmark  & \checkmark    &              & \checkmark   & \checkmark   & \checkmark \\
        ViSER$^*$~\cite{yang2021viser}                    &                   &                  &            &  \checkmark & \checkmark  & \checkmark  & \checkmark    &              & \checkmark   & \checkmark   & \checkmark \\
        BANMo$^*$~\cite{yang2022banmo}                    &                   & (\checkmark)$^3$ & (\checkmark)$^3$ &  \checkmark & \checkmark  & \checkmark  & \checkmark    &              & \checkmark   & \checkmark   & \checkmark \\
		\midrule
		Unsup3D~\cite{wu20unsupervised}     &                   &                  &            &             &             &             &               & \checkmark   &              & \checkmark   & \checkmark \\
		CSM~\cite{kulkarni19canonical}      &  \checkmark       &                  &            & \checkmark  &             &             &               &              &              & \checkmark   &         \\
        CMR~\cite{kanazawa18learning}       & (\checkmark)$^4$  & (\checkmark)$^4$ & \checkmark & \checkmark  &             &             & \checkmark    &              &              & \checkmark      & (\checkmark)$^2$ \\
        U-CMR~\cite{goel20shape}            &  \checkmark       &                  &            & \checkmark  &             &             & \checkmark    &              &              & \checkmark   & \checkmark       \\
        IMR~\cite{tulsiani2020imr}            &  \checkmark       &                  &            & \checkmark  &             &             & \checkmark    &              &              & \checkmark   & (\checkmark)$^2$       \\
        TTP~\cite{kokkinos2021point}            &  \checkmark       &                  &            & \checkmark  &             &             & \checkmark    &              &              & \checkmark   & (\checkmark)$^2$       \\
        UMR$^\dagger$~\cite{li20self-supervised}      &                   &                  &            & \checkmark  &             &             & \checkmark    &              &              & \checkmark   & (\checkmark)$^2$ \\
        VMR~\cite{li20online}     & (\checkmark)$^1$ & (\checkmark)$^4$ & \checkmark & \checkmark  &             &   & \checkmark    &              & \checkmark   & \checkmark   & (\checkmark)$^2$ \\
        \midrule
        Ours                                &                   &                  &            &  \checkmark & \checkmark  & \checkmark  & \checkmark    &              & \checkmark   & \checkmark   & \checkmark  \\
		\bottomrule
	\end{tabular}\label{table:related}
    \vspace{-0.1in}
\end{table*}

\section{Related Work}
We divide the vast literature of related work into two parts. The first one focuses on learning based approaches for 3D reconstruction with limited supervision. The second parts highlights related work for 3D reconstruction from images and video.

\subsection{Unsupervised and Weakly-supervised 3D Reconstruction}

A primary motivation for introducing machine learning in 3D reconstruction is to enable reconstruction from single views, which necessitates learning suitable shape priors.
In particular, we focus the discussion on unsupervised and weakly-supervised methods that do not require explicit 3D ground-truth for training.
Early unsupervised work include monocular depth predictors trained from egocentric videos of rigid scenes~\cite{garg16unsupervised,zhou17unsupervised}.

Others have explored weakly-supervised methods for learning full 3D meshes of object categories~\cite{kato2018renderer, kanazawa18learning, liu19soft, kato2019learning, wang2018pixel2mesh, henderson2019learning, goel20shape, li20self-supervised, li20online, wu2021derender, kokkinos2021point, Kokkinos_2021_CVPR}.
Many of these methods learn from still images and generally require masks and other additional supervision or prior assumptions, summarized in~\cref{table:related}.
In particular, \textbf{CMR}~\cite{kanazawa18learning} uses 2D keypoint annotations (in addition to masks) to initialize shape and viewpoint using Structure-from-Motion (SfM).
This is extended in the follow-up works in various ways.
\textbf{U-CMR}~\cite{goel20shape}, \textbf{TTP}~\cite{kokkinos2021point} and \textbf{IMR}~\cite{tulsiani2020imr} replace the keypoint annotations with a category-specific template shape beforehand.
With the template shape, extensive viewpoint sampling (camera multiplex) can be done to search for the best camera viewpoint for each training image~\cite{goel20shape}.
\textbf{UMR}~\cite{li20self-supervised} instead uses part segmentations from SCOPS~\cite{hung19scops:}, which also requires supervised ImageNet pretraining.
\textbf{VMR}~\cite{li20online} extends CMR with asymmetric deformation, and introduces a test-time adaptation procedure on individual videos by enforcing temporal consistency on the predictions produced by a pre-trained CMR model.
Note that we use videos to learn a 3D shape model \emph{from scratch}, whereas VMR starts with a pre-trained model and only performs online adaptation on videos.
\textbf{CSM}~\cite{kulkarni19canonical} and \textbf{articulated CSM}~\cite{kulkarni20articulation-aware} learn to pose an externally-provided (articulated) 3D template of an object category to images.
Unsup3D~\cite{wu20unsupervised} learns symmetric objects, like faces, without masks, but only with limited viewpoint variation.

Adversarial learning has also been explored to replace the need of multi-views for training~\cite{kudo18unsupervised, chen19unsupervised, henzler19escaping, nguyen-phuoc19hologan:, nguyen-phuoc20blockgan:, ye21shelf-supervised, schwarz20graf:, Niemeyer2020GIRAFFE, StyleGAN3D, chanmonteiro2020pi-GAN, pan2020gan2shape}.
The idea is to use a discriminator network to tell whether or not arbitrarily generated views of the learned 3D model are plausible, which provides signals to learn the geometry.
Although this approach does not require viewpoint annotations for individual images, it does rely on a reasonable approximation of the viewpoint distribution in the training data, from which random views are generated.
Overall, promising results can be achieved on synthetic data as well as a few real object categories, but general methods usually recover only coarse 3D shapes or 3D feature volumes that are difficult to extract.

\subsection{Reconstruction from Multiple Views and Videos}
Most works using multiple views and videos focus on reconstructing individual instances of an object.
Classic SfM methods~\cite{Faugeras01geometry, hartley04multiple} use multiple views of a rigid scene, with pipelines such as KinectFusion~\cite{newcombe11kinectfusion} and DynamicFusion~\cite{newcombe15dynamicfusion:} integrating depth sensors for reconstructing dense static and deformable surfaces.
Neural implicit surface representations have recently emerged for multi-view reconstruction~\cite{yariv2020multiview, wang2021neus, Oechsle2021ICCV}.
NeRF~\cite{mildenhall20nerf:} and its deformable extensions~\cite{park2021nerfies,gafni2020dynamic,tretschk2021nonrigid,raj2020pva,noguchi2021neural, pumarola20d-nerf:} synthesize novel views from densely sampled multi-views of a static or mildly dynamic scene using a Neural Radiance Field, from which explicit coarse 3D geometry can be further extracted.
A more recent line of work, such as LASR~\cite{yang21lasr:} and ViSER~\cite{yang2021viser}, optimizes a single 3D deformable model on an individual video sequence, using mask and optical flow supervision.
BANMo~\cite{yang2022banmo} further extends the pipeline and optimizes over a few video sequences of the same object instance, with the help of a pretrained DensePose~\cite{neverova20continuous} model.
However, these optimization-based models are typically sensitive to the quality of the sequences and tends to fail when only limited views are observed (see \cref{fig:lasr_compare}).
In contrast, by learning priors over a video dataset, our model can perform inference on a single image.

Other works that learn 3D categories form videos typically require some shape prior, such as a parametric shape model~\cite{loper15smpl:, bfm09}, and hence mostly focus on reconstruction of human bodies or faces~\cite{tung17self-supervised, arnab2019exploiting, doersch2019sim2real, kanazawa2019learning, zhang2019predicting, feng2018joint, tran2019learning, Kokkinos_2021_CVPR, Zuffi19Safari}.
{}\cite{novotny17learning} and~\cite{henzler21unsupervised} consider turn-table like videos to learn to reconstruct rigid object categories.
In contrast, our method learns a 3D shape model of a deformable object category \emph{from scratch} using  videos.
\begin{figure*}[t]
    \centering
    \includegraphics[width=\linewidth]{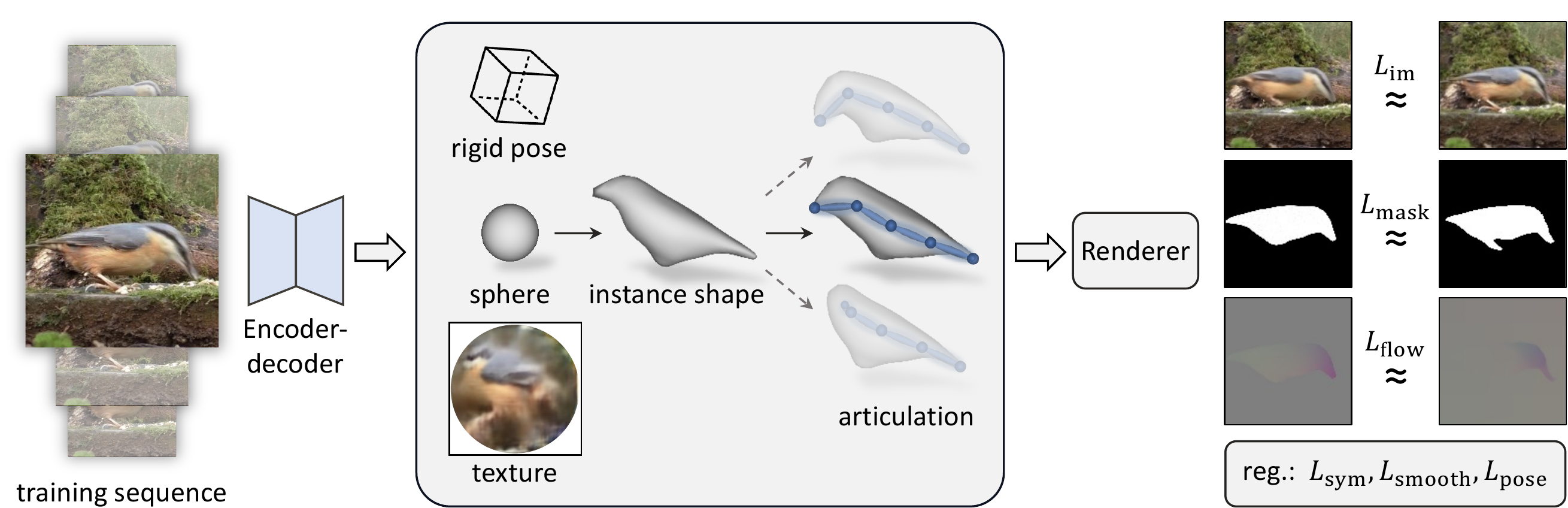}
    \vspace{0.05in}
    \caption{\textbf{Training Pipeline.} Form a single frame of a video, we predict the 3D pose, shape and texture of the object. The shape is further disentangled into category shape, instance shape and deformation using linear blend skinning. Using a differentiable rendering step, we can train the model end-to-end by reconstructing the image and by enforcing temporal consistencies.}
    \label{fig:pipeline}
\end{figure*}

\section{Method}

Our goal is to learn a function $(V,\xi,T)=f(I)$ that, given a \emph{single image} $I\in\mathbb{R}^{3\times H \times W}$ of an object, predicts its 3D shape $V$ (a mesh), its pose $\xi$ (either a rigid transformation or full articulation parameters) and its texture $T$ (an image).
We describe below the key ideas in our method and refer the reader to the sup.~mat.~for details.

While the predictor $f$ is monocular, we supervise it by making use of video sequences $\mathcal{I}=\{I_t\}_{t=1,\dots,|\mathcal{I}|}$, where $t$ denotes time.
For this, we use a \emph{photo-geometric auto-encoding approach}.
Let $M \in \{0,1\}^{H\times W}$ be the 2D mask of the object in image $I$, which we assume to be given.
The model $(V,\xi,T)=f(I)$ encodes the image as a set of photo-geometric parameters; from these, an handcrafted rendering function $(\hat I,\hat M) = \mathcal{R}(V,\xi,T)$ reconstructs the image $\hat I$ and the mask $\hat M$.
For supervision, the rendered image and the rendered mask is compared to the given ones via two losses:
\begin{align}
L_{\text{im}} &= \lambda_\text{im}\|\hat M \odot (\hat{I} - I)\|_1, \\ L_{\text{mask}} &= \lambda_\text{mask}\|\hat M - M\|_2^2 \, ,
\end{align}
where $ \lambda_\text{im}$ and $ \lambda_\text{mask}$ weigh each loss.
Note that the image loss is restricted to the predicted region as the model only represents the object but not the background.
\Cref{fig:pipeline} gives an overview of the training pipeline.

\subsection{Solving the Viewpoint Ambiguity}\label{s:ambiguities}

Decomposing a single image into shape, pose and appearance is ambiguous, 
which is a major challenge that can easily result in poor reconstructions.
Some prior works have addressed this issue by sampling a large number of viewpoints during training, thus giving the optimizer a chance to avoid local optima.
However, this is a slow process that requires testing a large number of hypotheses at every iteration (\eg 40 in~\cite{goel20shape}) and requires a precise template shape to understand the differences between small viewpoint changes.

Here we note that this is likely unnecessary.
The key observation is that the ambiguities arising from image-based reconstruction are not arbitrary; instead, they tend to concentrate around specific symmetries induced by the projection of a 3D object onto an image.

\setlength{\intextsep}{6pt}
\setlength{\columnsep}{12pt}
\begin{wrapfigure}{r}{0.5\columnwidth}
\includegraphics[width=0.5\columnwidth]{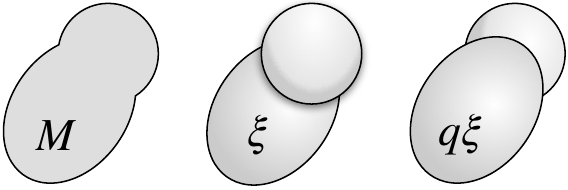}
\end{wrapfigure}
The image to the right illustrates this idea.
Here, given only the mask $M$, one is unable to choose between the object pose $\xi$ or its mirrored variant $q\xi$, where $q$ is a suitable `mirror mapping' that rotates the pose back to front (see sup.~mat.~for details).
We argue that, before developing a more nuanced understanding of appearance, the model $f$ is similarly undecided about the pose of the 3D object; however, the number of choices is very limited: either the current prediction $\xi = f_\xi(I)$ is correct, or its mirrored version $q \xi$ is.

Concretely, during training we evaluate the loss $L(V,\xi,T)$ for the model prediction and the loss $L(V,q\xi,T)$ for the mirrored pose.
We find the better of the two poses $\xi^* = \argmin_{\hat{\xi} \in \{\xi, q\xi \}} L(V,\hat{\xi},T)$ and optimize the loss:
\begin{equation}
    L_\text{pose} = \lambda_\text{pose}\|\xi - \xi^* \|^2_2 \, .
\end{equation}
In this way, the model is encouraged to flip its prediction when $L(V,q\xi,T) < L(V,\xi,T)$.
This assures that the model eventually learns the correct pose and does not rely on the flipping towards the end of the training.

\subsection{Learning from Videos}

We exploit the information in videos by noting that the shape $V$ and texture $T$ of an object are invariant over time, with any time-dependent change limited to the pose $\xi$.
Hence, given a sequence of images $\mathcal{I}=\{I_t\}_{t=0,\dots,|\mathcal{I}|}$ of the same object and corresponding frame-based predictions $(V_t,\xi_t,T_t) = \Phi(I_t)$, we feed the rendering function $(\hat I_t, \hat M_t) = R(\bar V, \xi_t, \bar T)$ with the shape and texture averages 
$
\bar V = \frac{1}{|\mathcal{I}|} \sum_{t=1}^{|\mathcal{I}|} V_t
$
and 
$
\bar T = \frac{1}{|\mathcal{I}|} \sum_{t=1}^{|\mathcal{I}|} T_t.
$
The idea is that, unless shape and texture agree across predictions, their averages would be blurry and result in poor renderings.
Hence, minimizing the rendering loss indirectly encourages these quantities to be consistent over time.

Furthermore, while the pose $\xi_t$ does vary over time, pose changes must be compatible with image-level correspondences.
Specifically, let $F_t \in \mathbb{R}^{H \times W \times 2}$ be the optical flow measured between frames $I_t$ and $I_{t+1}$ by an off-the-shelf method such as RAFT~\cite{teed20raft:}.
We can render the flow $\hat F_t = \mathcal{R}(V,\xi_t,\xi_{t+1})$ by computing the displacement of the object vertices $V$ as a pose change from $\xi_t$ to $\xi_{t+1}$.
We can then add the flow reconstruction loss
\begin{equation}
L_{\text{flow}}(\hat F_t, F_{t}) =  \lambda_\text{flow}
\|
M_t \odot (\hat{F}_{t}- F_{t})
\|_2^2 \, ,
\end{equation}
to encourage consistent motion of the object. Its influence is controlled by the weight $\lambda_\text{flow}$.

\subsection{Hierarchical Shape Model}

Next, we flesh out the shape model.
The shape $V \in \mathbb{R}^{3 \times K}$ is given by $K$ mesh vertices and represents the shape of a specific object instance in a \emph{canonical} pose.
It is obtained by the predictor
$
f_V(I) = V_\text{base} + \Delta V_\text{tmpl} + \Delta V(I)
$
where:
$V_\text{base}$ is an initial fixed shape (a sphere),
$\Delta V_\text{tmpl}$ is a learnable matrix (initialized as zeros) such that $V_\text{tmpl} = V_\text{base} + \Delta V_\text{tmpl}$ gives an average shape for the category (template), and $\Delta V(I)$ is a neural network further deforming the this template into the specific shape of the object seen in image $I$.
We further restrict $V$, which is the rest pose, to be bilaterally symmetric by only predicting half of the vertices and obtaining the remaining half via mirroring along the $x$ axis.
Note that, while in many prior works the category-level template $V_\text{tmpl}$ is given to the algorithm, here this is learned automatically from a sphere.

Finally, the shape $V$ is transformed into the actual mesh observed in the image by a \emph{posing function}
$
g(V, \xi).
$
We work with two kinds of such functions.
The first one is a simple rigid motion
$
g(V,\xi) = g_\xi V,
$
$
g_\xi \in SE(3).
$
This is used in an initial warm-up phase for the model to allow it to learn a first version of the template $V$ automatically.

In a second learning phase, we further enrich the model to capture  complex articulations of the shape.
There are a number of possible parameterizations that could be used for this purpose.
For instance,~\cite{Kokkinos_2021_CVPR} automatically initializes a set of keypoints via spectral analysis of the mesh.
Here, we initialize instead a traditional skinning model, given by a system of bones $b\in \{1,\dots, B\}$, ensuring inelastic deformations.

The skinning model is specified by: the bone topology (a tree), the joint location $\mathbf{J}_b \in\mathbb{R}^{3}$ of each bone with respect to the parent bone, the relative rotation $\xi_b \in SO(3)$ of that bone with respect to the parent, and a row-stochastic matrix of weights $w \in [0,1]^{K\times B}$ specifying the strength of association of each mesh vertex to each bone.
Of these, only the topology is chosen manually (e.g, to account for a different number of legs for objects in the category). The joint locations $\mathbf{J}_b$ and the skinning weights $w$ are set automatically based on a simple heuristic (described in sup.~mat.).

While topology, $\mathbf{J}_b$ and $w$ are fixed, the joint rotation $\xi_b \in SO(3), b=2,\dots,B$ and the rigid pose $\xi_1 \in SE(3)$ are output by the predictor $f$ to express the deformation of the object as it changes from image to image.

\subsection{Appearance Model and Rendering}

We model the appearance of the object using a texture map $T \in \mathbb{R}^{3\times H_T \times W_T}$.
The vertices of the base mesh $V_\text{base}$ are assigned to fixed texture uv-coordinates and the texture inherits the symmetry of the base mesh.
Given the posed mesh $g(V,\xi)$ and the texture $T$, we \emph{render} an image $(\hat I, \hat M)=\mathcal{R}(V,\xi,T)$ of the object using standard perspective-correct texture mapping with barycentric coordinates using the PyTorch3D differentiable mesh renderer~\cite{ravi20accelerating}.

\subsection{Symmetry and Geometric Regularizers}

An important property of object categories is that they are often symmetric.
This does not mean that individual object instances are symmetric, but that the space of objects is~\cite{thewlis18modelling}.
In other words, if image $I$ contains a valid object, so does the mirrored image $m I$.
Furthermore, given the photo-geometric parameters $(V,\xi,T) = f(I)$ for $I$, the parameters for $m I$ must be given by $f(m I) = (m V, m \xi, Tm)$ where $m V = V$ (because the rest shape is assumed symmetric), $Tm$ is the flipped texture image and $m \xi$ is a mirrored version of the pose.
Hence, we additionally enforce the pose predictor to satisfy this structure by minimizing the loss
$
L_{\text{sym}} =  \lambda_\text{sym}\|f_\xi(m I) - m(f_\xi(I))\|_2^2\, ,
$
weighted by $\lambda_\text{sym}$.

Note the relationship between the mirroring operators $q\xi$ in~\cref{s:ambiguities} and $m\xi$ here: they are the same, up to a further rigid body rotation.
The effect is that $q$ appears to rotate the object back to front, and $m$ left to right.
This is developed formally in the sup.~mat.

We further regularise learning the mesh $V$ via a loss 
$
L_\text{smooth}(V,V_\text{tmpl})
$
which includes:
the ARAP loss~\cite{sorkine2007rigid} between $V$ and the template $V_\text{tmlp}$, ensuring that they do not diverge too much, and a Laplacian and mesh normal smoothers for $V$.

\subsection{Learning Formulation}\label{sec:learning}

Given a video $\mathcal{I} = \{I_t\}_{t=1,\dots,|\mathcal{I}|}$, the overall learning loss is thus:
$$
L
=
L_\text{im} +
L_\text{mask} +
L_\text{flow} +
L_\text{sym} +
L_\text{smooth} + L_\text{pose}.
$$
In practice, we found it important to warm up the model, activating increasingly more refined model components as training progresses.
This can be seen as a sort of coarse-to-fine or paced learning strategy.

Learning thus uses the following schedule in three phases:
(1) \emph{shape learning}: the basic model with no instance-specific deformation (i.e., $V=V_\text{tmpl}$), no bone articulation and only the mask loss is optimized in order to obtain an initial template $V_\text{tmpl}$;
(2) \emph{ambiguity resolution}: the pose rectification loss $L_\text{pose}$ is activated to resolve the front-to-back ambiguity of~\cref{s:ambiguities};
(3) \emph{full model}: the bones are instantiated, and the instance deformation, skinning models and appearance loss are also activated in order to learn the full model.
\section{Experiments}

\begin{figure}[t]
    \includegraphics[width=1\linewidth]{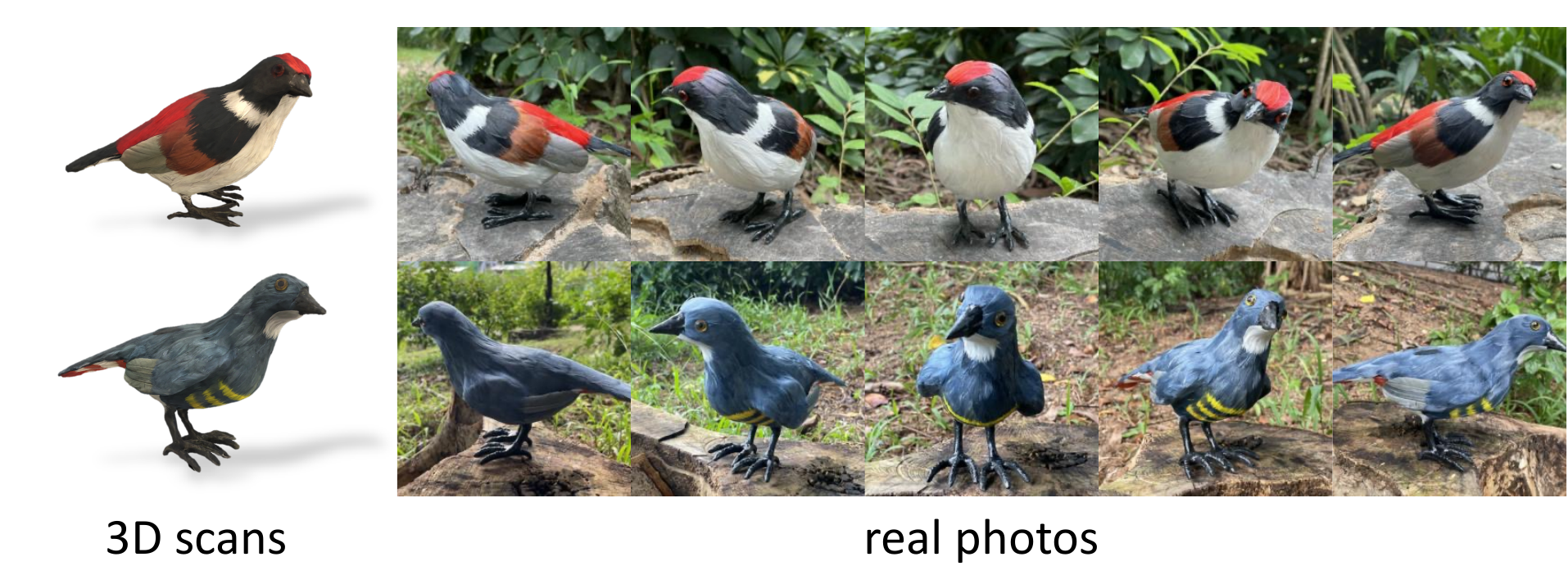}
    \caption{\textbf{Examples of the 3D Toy Bird Dataset.}
    Each bird toy was 3D scanned and the photographed ``\textit{in the wild}''.}
    \label{fig:toy_bird_sample}
\end{figure}

\begin{figure*}[t]
    \centering
    \includegraphics[trim={0 0 20px 0}, clip, width=\linewidth]{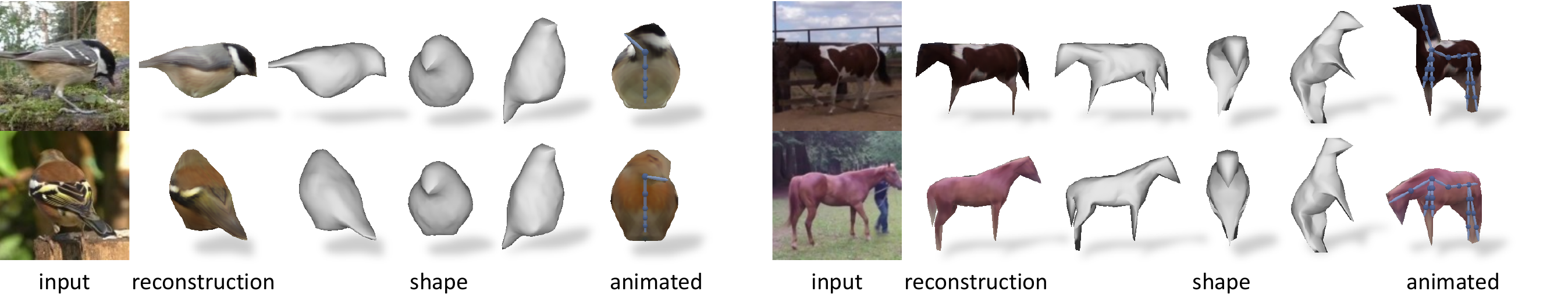}
    \caption{\textbf{Qualitative Examples.} We show multiple views of the reconstructed mesh together with a textured view and animated version of the bird that we obtained by rotating the learned bones. We find that the model is able to recover the shape well even when seen from novel viewpoints. The animation is able to generate believable poses.}
    \label{fig:recon}
\end{figure*}

We perform an extensive set of experiments to evaluate our method and compare to the state of the art. To this end, we collect three datasets, two of real animals, birds and horses, and one using toy birds of which we can obtain ground truth 3D scans.

\subsection{Dataset and Implementation Details}\label{sec:dataset}

\paragraph{Video Datasets.}
We experiment with two types of objects: birds and horses.
For each category, we extract a collection of short video clips from YouTube.
The exact links to these videos and the preprocessing details are included in the sup.~mat.
We use the off-the-shelf PointRend model~\cite{kirillov2019pointrend} to detect and segment the object instances, remove the frames where the object is static, and
automatically split the remaining frames into short clips, each containing one single object.
The frames and the masks are then cropped around the objects and resized to $128 \times 128$ for training.
We also run the off-the-shelf RAFT model~\cite{teed20raft:} on the full frames to estimate optical flow between consecutive frames, and account for the cropping and resizing to obtain the correct optical flow for the crops.
This procedure creates $1,962$ and $114$ short clips of birds and horses respectively, each containing $16$ to a few hundred frames with paired image, mask and flow.
We randomly split them into $1,767$/$195$ and $103$/$11$ training/testing sequences for birds and horses respectively.

\begin{table*} [t!]
    \small
    \newcommand{\xpm}[1]{{\tiny$\pm#1$}}
	\centering
    \caption{\textbf{Evaluation on Toy Bird Scans.} Shape reconstruction quality measured by bi-directional Chamfer Distance between predicted shape and ground-truth scans. The lower the better. \stemp template shape, \sview viewpoint, \skey 2D keypoint, \smask mask, \sflow optical flow, \svid video.
    ``finetuned'' indicates pretrained models finetuned on our video dataset.}
	\begin{tabular}{lcc}
		\toprule
		 & Supervision & Chamfer Distance (cm) $\downarrow$  \\
		\midrule
		 CMR~\cite{kanazawa18learning} (finetuned) & \supcmr   & 1.35 \xpm{0.81} \\
		 U-CMR~\cite{goel20shape} (finetuned) & \supucmr & 1.82 \xpm{0.93} \\
		 VMR~\cite{li20online} (finetuned) & \supvmr  & 1.28 \xpm{0.69} \\
		 UMR~\cite{li20self-supervised} (finetuned) & \supumr+SCOPS   & 1.24 \xpm{0.75}  \\
		 \midrule
		 CMR~\cite{kanazawa18learning} & \smask   & 5.94 \xpm{10.33} \\
		 U-CMR~\cite{goel20shape} & \smask & 4.36 \xpm{1.56} \\
		 VMR~\cite{li20online} & \smask  & 1.90 \xpm{0.96} \\
		 UMR~\cite{li20self-supervised} & \smask   & 2.26 \xpm{1.12}  \\
		 UMR~\cite{li20self-supervised} & \smask+SCOPS   & 1.82 \xpm{0.93}  \\
		 \midrule
		 Ours & \supours & \textbf{1.51} \xpm{0.89}  \\ 
		\bottomrule
	\end{tabular}
	\label{tab:toy_bird}
\end{table*}

\paragraph{3D Toy Bird Dataset.}
In order to properly evaluate and compare the quality of the reconstructed 3D shapes produced by different methods, we introduce a 3D Toy Bird Dataset, which consists of ground-truth 3D scans of realistic toy bird models and photographs of them taken in real world environments.
\cref{fig:toy_bird_sample} shows examples of the dataset.
Specifically, we obtain $23$ toy bird models, and used Apple RealityKit Object Capture API~\cite{AppleObjectCapture} to capture accurate 3D scans from turn-table videos.
For each model, we then take $5$ photographs from different viewpoints in $3$ different outdoor scenes, resulting in a total of $345$ images.
We will release the dataset and ground-truth for future benchmarking.

\paragraph{Implementation Details.}\label{sec:impl-details}

Our reconstruction model is implemented using three neural networks ($f_V$, $f_\xi$, $f_T$) as well as a set of of trainable parameters for the categorical prior shape $\Delta V_\text{tmpl}$.
The shape network $f_V$ and the rigid pose network $f_\xi$ are simple encoders with downsampling convolutional layers that take in an image and predict vertex deformations $\Delta V_\text{ins}$, skinning parameters $\xi_{2:B}$, and rigid pose $\xi_1$ and $\mathbf{J}_1$ as flattened vectors.
The texture network $f_T$ is an encoder-decoder that predicts the texture map $T$ from an image.
We use Adam optimizers with a learning rate of $10^{-4}$ for all networks, and a learning rate $0.01$ for the category shape parameters $\Delta V_\text{tmpl}$.
We use a symmetric ico-sphere 
as the initial mesh.
For each training iteration, we randomly sample $8$ consecutive frames from $8$ sequences.
The models are trained in three phases described in~\cref{sec:learning}.
All details are included in the sup.~mat.

\begin{table*} [t!]
    \small
    \newcommand{\xpm}[1]{{\tiny$\pm#1$}}
	\centering
  \setlength{\tabcolsep}{3.5pt}
    \caption{\textbf{Mask Forward Projection IoU.} Shape reconstruction quality and temporal consistency measured by projecting the shape predicted at frame $t$ to a different pose at a \emph{future} frame $t + \Delta t$ and comparing the masks at $t + \Delta t$. The higher the better. ``finetuned'' indicates pretrained models finetuned on our video dataset.
    }
	\begin{tabular}{lcccc}
		\toprule
		Frame offset & Supervision & $\Delta t= 0$ & $\Delta t= 5$ & $\Delta t= 20$  \\
		\midrule
        CMR~\cite{kanazawa18learning} (finetuned) & \supcmr & 0.770 \xpm{0.13} & 0.722 \xpm{0.13} & 0.712 \xpm{0.13} \\
        U-CMR~\cite{goel20shape} (finetuned) & \supucmr & 0.790 \xpm{0.06} & 0.761 \xpm{0.07} & 0.758 \xpm{0.07} \\
        VMR~\cite{li20online} (finetuned) & \supvmr & 0.807 \xpm{0.08} & 0.752 \xpm{0.08} & 0.737 \xpm{0.09} \\
        UMR~\cite{li20self-supervised} (finetuned) & \supumr+ SCOPS & 0.847 \xpm{0.05} & 0.782 \xpm{0.07} & 0.772 \xpm{0.07} \\
        \midrule
        CMR~\cite{kanazawa18learning} & \smask & 0.634 \xpm{0.06} & 0.605 \xpm{0.11} & 0.596 \xpm{0.11} \\
        U-CMR~\cite{goel20shape} & \smask & 0.725 \xpm{0.05} & 0.714 \xpm{0.06} & 0.700 \xpm{0.06} \\
        VMR~\cite{li20online} & \smask & 0.777 \xpm{0.05} & 0.720 \xpm{0.07} & 0.700 \xpm{0.09} \\
        UMR~\cite{li20self-supervised} & \smask & 0.853 \xpm{0.04} & 0.798 \xpm{0.06} & 0.788 \xpm{0.06} \\
        UMR~\cite{li20self-supervised} & \smask+ SCOPS & 0.830 \xpm{0.04} & 0.766 \xpm{0.07} & 0.753 \xpm{0.07} \\
		 \midrule

		 Ours (articulation fixed) & \supours & \textbf{0.855} \xpm{0.07} & 0.810 \xpm{0.07} & 0.805 \xpm{0.07} \\ 
		 Ours (articulation transferred) & \supours  & \textbf{0.855} \xpm{0.07} & \textbf{0.844} \xpm{0.07} & \textbf{0.845} \xpm{0.07} \\ 
		\bottomrule
	\end{tabular}
	\label{table:miou}
\end{table*}

\begin{figure*}[t]
    \centering
    \includegraphics[width=\linewidth]{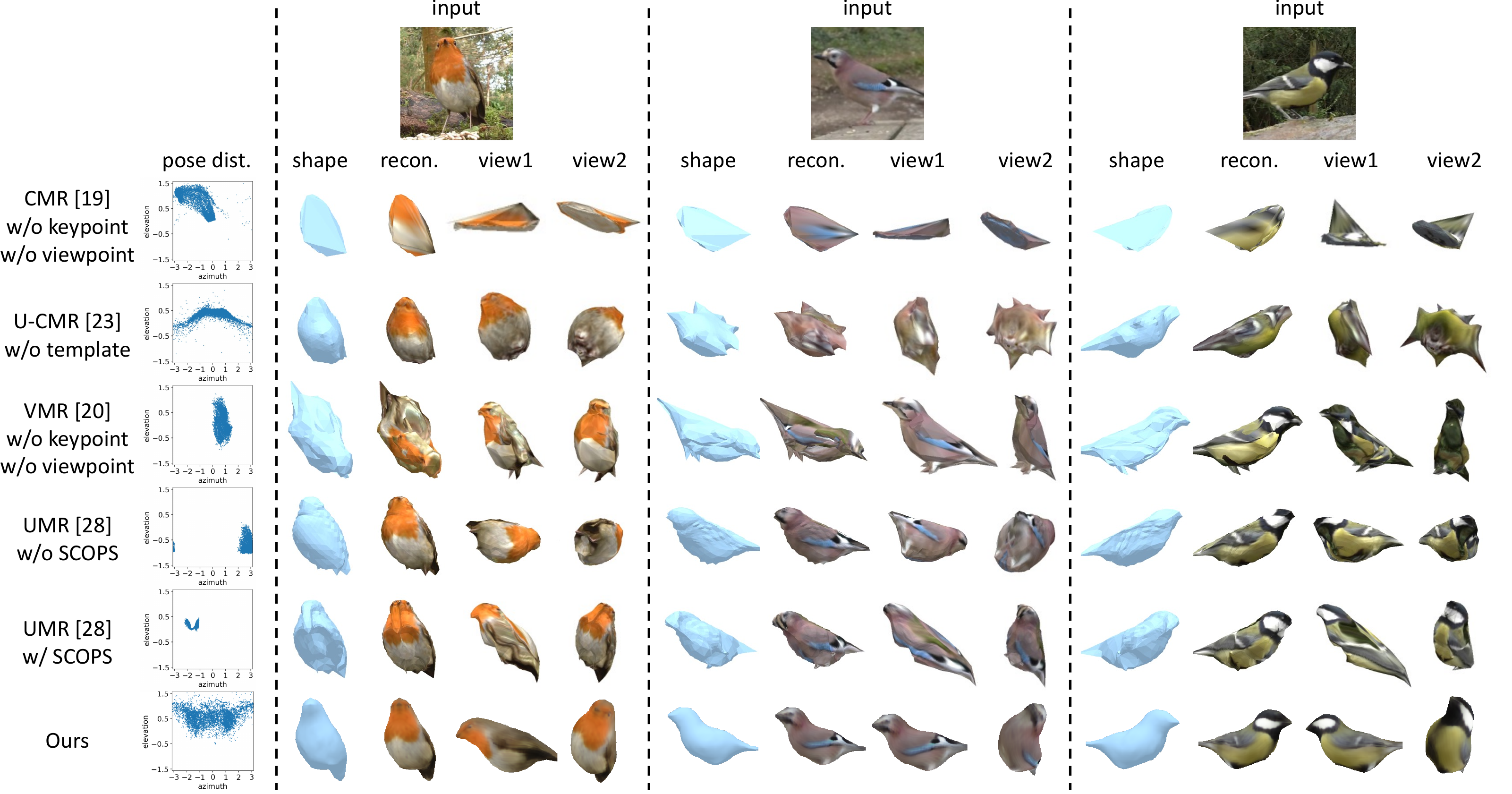}
    \caption{\textbf{Visual Comparison.} We compare to state-of-the-art methods trained without external geometric supervision in the form of 2D keypoints, viewpoint, or template shape. As UMR leverages weak-supervision using part segmentation maps from SCOPS~\cite{hung19scops:}, we show versions trained with and without SCOPS. 
    Our method consistently reconstructs reasonable 3D shapes and the predictions cover full 360-degree (azimuth) view, whereas other methods produce poor reconstructions and their viewpoint predictions collapse to only a limited range with the exception of U-CMR.
    Other methods, except for U-CMR directly copy the texture from the input image using texture flow. Hence, although the texture appears sharper from the input view, they are often incorrect as seen from other views.
    See the sup.~mat. for extended results. %
    }
    \label{fig:vis_compare}
\end{figure*}

\subsection{Qualitative Results}

\Cref{fig:recon} shows qualitative 3D reconstruction results obtained from our model.
Note that videos are no longer needed during inference and the shown predictions come from a single frame.
Despite not requiring any explicit 3D, viewpoint or keypoint supervision, our model learns to reconstruct accurate 3D shapes from only monocular training videos.
The reconstructed 3D meshes can be animated with our skinning model by transforming the bones of the learned shape. This animation can also be transferred between instances.

\subsection{Comparisons with State-of-the-Art Methods}

We compare our model with a number of state-of-the-art learning-based reconstruction methods, including CMR~\cite{kanazawa18learning}, U-CMR~\cite{goel20shape}, UMR~\cite{li20self-supervised} and VMR~\cite{li20online}.
CMR requires 2D keypoint annotations for initializing the 3D shape and viewpoints and also for the training loss.
U-CMR removes keypoint supervision but requires a 3D template shape, and UMR replaces that with part segmentation maps from SCOPS~\cite{hung19scops:} which relies on supervised ImageNet pretraining.
VMR~\cite{li20online} allows for deformations but it requires the same level of supervision as CMR\@.
All of them rely on external geometric supervision to establish correspondences for learning 3D shapes.
We train all these methods on our video dataset with only mask supervision and show that without the additional supervision, all these methods reconstruct poor shapes.
We also finetune their models pre-trained on CUB~\cite{WahCUB_200_2011} with the required keypoint, camera view or template shape supervision on our bird video dataset.
Finally, we also train UMR from scratch on our bird video dataset with SCOPS predictions obtained from the pre-trained SCOPS model.

\paragraph{On 3D Toy Bird Scans.}
Our toy scan dataset allows for a direct evaluation of shape prediction.
We first scale the predicted shapes to match the volume of the scans and roughly align the canonical pose of each method to the scans manually.
Each individual predicted shape is further aligned to the ground-truth scan using Iterative Closest Point (ICP)~\cite{besl92a-method} and the symmetric (average of scan-to-object and object-to-scan) Chamfer distance is reported in centimeters, \cref{tab:toy_bird}, by assuming the width of each bird to be 10 cm.
While the reconstruction quality of other methods is good when trained with more geometric supervision, it degrades strongly without this training signal resulting in worse reconstructions when compared to our method.
Note that this metric evaluates the individual shape predictions regardless the viewpoints.
Next we evaluate the consistency across views.

\paragraph{On Bird Video Dataset.}

Since we do not have ground-truth 3D shape and viewpoints for direct evaluation on our video test set, we measure reconstruction quality via a mask \emph{forward} projection accuracy from one frame to another, using the object masks predicted by PointRend~\cite{kirillov2019pointrend} as the pseudo ground-truth.
This evaluates the shape and viewpoint quality as the object from a past frame is projected to a future frame which can only align when both shape and pose are estimated correctly, but cannot account for non-rigid deformation between frames.
For each test sequence, we predict the shape at frame $t$ and render the object mask from the pose at frame $t + \Delta t$ with an offset $\Delta t$ of 0, 5 and 20 frames.
We then compute the mean Intersection over Union (mIoU) between the rendered masks and the ground-truth masks at $t + \Delta t$.
\cref{table:miou} summarizes the results, which suggest that our model achieves both better shape reconstruction and viewpoint consistency.
We also compute the metrics on our model with frame-specific deformations predicted at frame $t + \Delta t$ applied to the shape predicted at frame $t$.
This further improves the mask reprojection IoU, confirming that our model learns correct frame-specific deformations.
Other methods tend to overfit the shape to the image, resulting in a larger decrease in reprojection accuracy with increasing $\Delta t$.

We also compare the distribution of estimated viewpoints/object poses by plotting the elevation and azimuth predicted on the test set in~\cref{fig:vis_compare}.
Our method is able to learn the full azimuth range, while other methods, with the exception of U-CMR, only predict limited range of views (azimuth) without additional geometric supervision.

\begin{figure}[t]
    \includegraphics[width=1\linewidth]{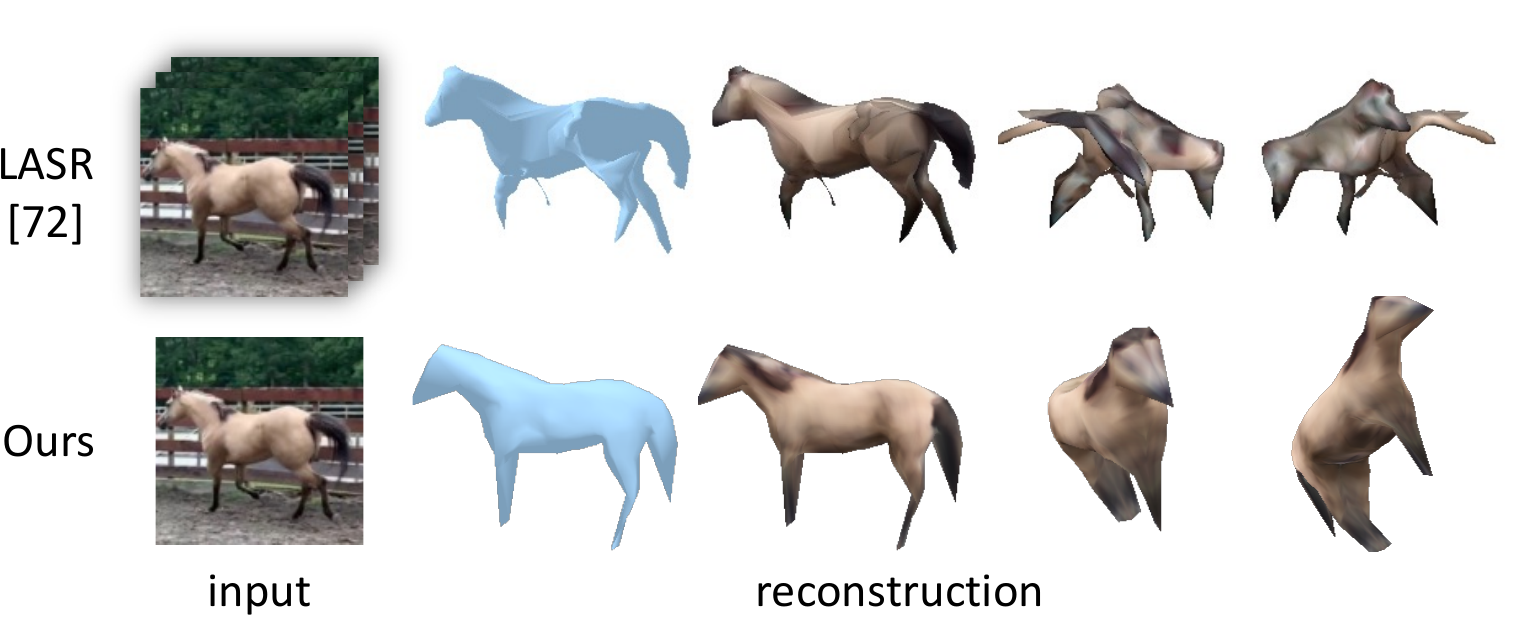}
        \caption{\textbf{Comparison with LASR~\cite{yang21lasr:}.} While the rendering in the original viewpoint looks convincing, the shape produced by LASR is distorted and does not resemble the actual shape of a horse. Since our method trains on multiple sequences it can learn a consistent shape.}
    \label{fig:lasr_compare}
\end{figure}

\begin{figure*}[t]
  \centering
  \captionsetup[subfigure]{justification=centering,labelsep=space,aboveskip=2pt}
  \includegraphics[trim={0 0 20px 0}, clip, width=\linewidth]{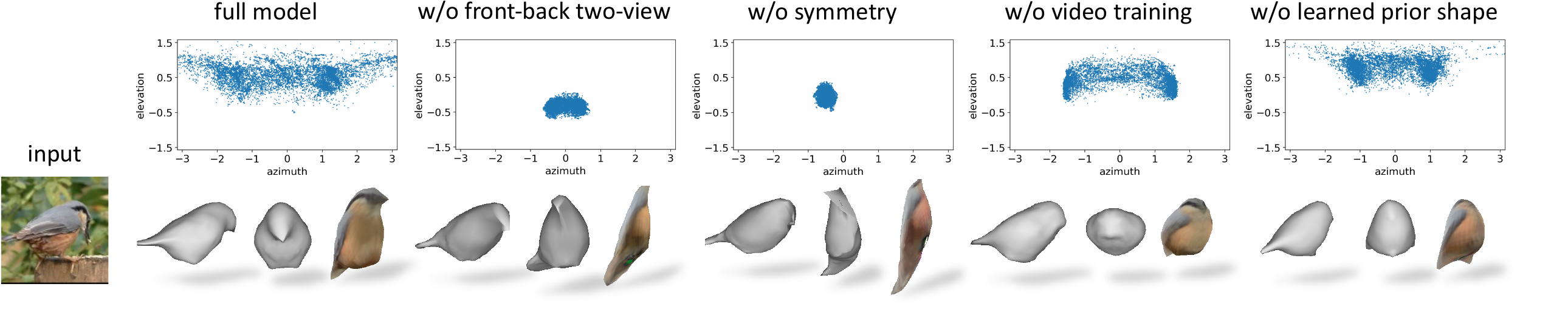}
  \vspace{-0.1in}
\caption{\textbf{Ablation Studies.} We train our model without some of the key components and plot the distribution of the predicted poses. Without 2-view ambiguity resolution or symmetry constraint, the pose prediction collapses. Video training and learning a shape prior also help improve the poses and shapes.}
\label{fig:ablation}
\end{figure*}

\paragraph{Qualitative Comparisons.}

\cref{fig:vis_compare} shows a qualitative comparison of different methods.
When methods relying on more geometric supervision (CMR, U-CMR, VMR) are trained without this learning signal, they fail to produce reasonable shape reconstructions.
UMR trained without SCOPS part segmentations overfits to the input views producing inaccurate 3D shapes.  
Our method reconstructs accurate shape and pose, despite not using keypoint or template supervision.
We refer the reader to the sup.~mat.~for more results.
Note that our model is trained on $128 \times 128$ images, whereas other methods train on $256 \times 256$ images and, except U-CMR, sample the texture directly from the input image, explaining the difference in the texture quality. %

\begin{table} [t!]
    \small
    \newcommand{\xpm}[1]{{\tiny$\pm#1$}}
	\centering
    \caption{\textbf{Ablation Studies with 3D Toy Bird Scans.} Every component of our model helps to improve the final performance.}
	\begin{tabular}{lcc}
		\toprule
		 & Chamfer Distance (cm)  \\
		\midrule
		 Full model & \textbf{1.51} \xpm{0.89} \\
		 \midrule
		 w/o front-back hypothesis & 2.52 \xpm{1.41} \\
		 w/o symmetry & 2.19 \xpm{1.24} \\
		 w/o video training  & 2.20 \xpm{1.03} \\
		 w/o learned prior shape & 3.92 \xpm{1.47} \\
		\bottomrule
	\end{tabular}
	\label{table:abl_3d}
\end{table}

\paragraph{On Horse Video Dataset.}
For horses, we compare qualitatively with LASR~\cite{yang21lasr:} in \cref{fig:lasr_compare}, which is an optimization-based method for single video sequences. While their reconstruction appears to be  convincing in the original viewpoint, the actual mesh often does not resemble the shape of a horse.
Running LASR on such a sequence takes over four hours.

\subsection{Ablation and Analysis}

We ablate the different components of our method quantitatively on our toy bird dataset in \cref{table:abl_3d} and \cref{fig:ablation}. We find that all components are necessary for the final performance. The pose distribution in \cref{table:abl_3d} shows that the model only learns the full 360-degree (azimuth) view of the object when all components are active. Especially the two-view-ambiguity resolution and the shape symmetry are important to learn the pose while video training helps to discover the backside of the object. Without a good pose prediction the reconstructions look reasonable in the input view, reveal to be degenerate from other directions.

The model without symmetry produces unrealistic shapes indicating that symmetry is a useful prior, even when learning deformable shapes.
Similarly, the shape prior is important to discover fine details (\eg beak and tail) that are not visible in every image.
The full model predicts a full range of viewpoints (\cref{fig:ablation}) and the most consistent shape (\cref{table:abl_3d}).

We train another model without the learned category prior shape, predicting individual shapes for each bird.
The resulting reconstructions are inconsistent across different instances, shown in~\cref{fig:ablation}.
This suggests that the full model is able to leverage shape prior of the whole category, which is a major benefit of \textit{learning} in a reconstruction pipeline.

\section{Limitations and Future Work}\label{sec:limitations}

Our method still requires segmentation masks obtained from the off-the shelf model as supervision for training. 
Moreover, their quality affects the fidelity of our reconstructions.
Thus, similar to comparable methods,
our reconstructions do not capture fine details well, such as legs and the beak.
The texture prediction sometimes results in low quality reconstructions especially when the input image is affected by motion blur.
Currently, we have to handcraft a structure for various types of animals, for example different structures for horses (quadrupeds) and birds.
How to automatically discover plausible bone structures is also an interesting question to explore for future work.
\section{Conclusions}
We have presented a method to learn articulated 3D representations of deformable objects from monocular videos without explicit geometric supervision, such as keypoints, viewpoint or template shapes.
The resulting 3D meshes are temporally consistent and can be animated.
The method can be trained from videos and only needs off-the-shelf object detection and optical flow models for preprocessing.
For reproducibility, comparison and benchmarking, the dataset, code and models will be released with the paper.

\section*{Acknowledgments}
We thank Zirui Wang for insightful discussions and Xueting Li for sharing the code for VMR with us. Shangzhe Wu is supported by Meta Research. Tomas Jakab is supported by Clarendon Scholarship. Christian Rupprecht is supported by Innovate UK (project 71653) on behalf of UK Research and Innovation (UKRI) and the Department of Engineering Science at the University of Oxford. Andrea Vedaldi is supported by EPSRC Grant (Visual AI EP/T028572/1).

\bibliography{ref}%

\clearpage
\onecolumn

\section{Appendices}

\subsection{Additional Results}\label{s:sup_res}
On top of the material in this pdf, please see the \textbf{supplementary video} for more qualitative results.

\subsubsection{Additional Comparisons with State-of-the-Art Methods}\label{s:sup_compare}

\begin{figure}[t!]
    \centering
    \includegraphics[width=\linewidth]{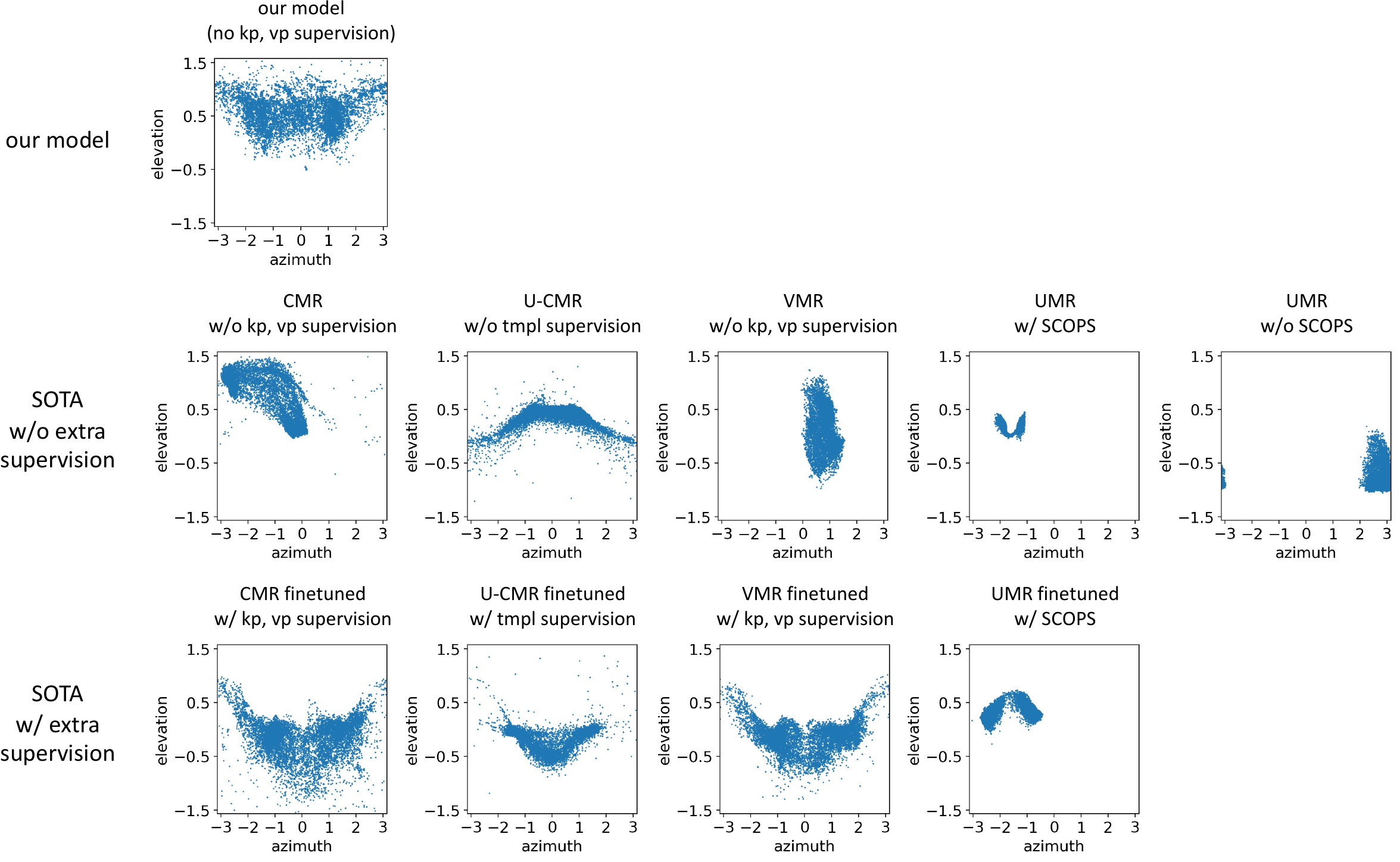}
    \caption{\textbf{Comparison of the Learned Rigid Pose Distributions.} We plot the distributions of the rigid poses predicted by various methods. Without additional geometric supervision (``kp'' for keypoint, ``vp'' for viewpoint and ``tmpl'' for template shape), the rigid poses (viewpoint) predicted by existing methods collapse to only a limited range, hence resulting in poor, inconsistent shape predictions. For example, UMR is able to predict only frontal poses.}
    \label{fig:supmat_pose_compare}
\end{figure}

\begin{figure}[t!]
    \centering
    \includegraphics[width=\linewidth]{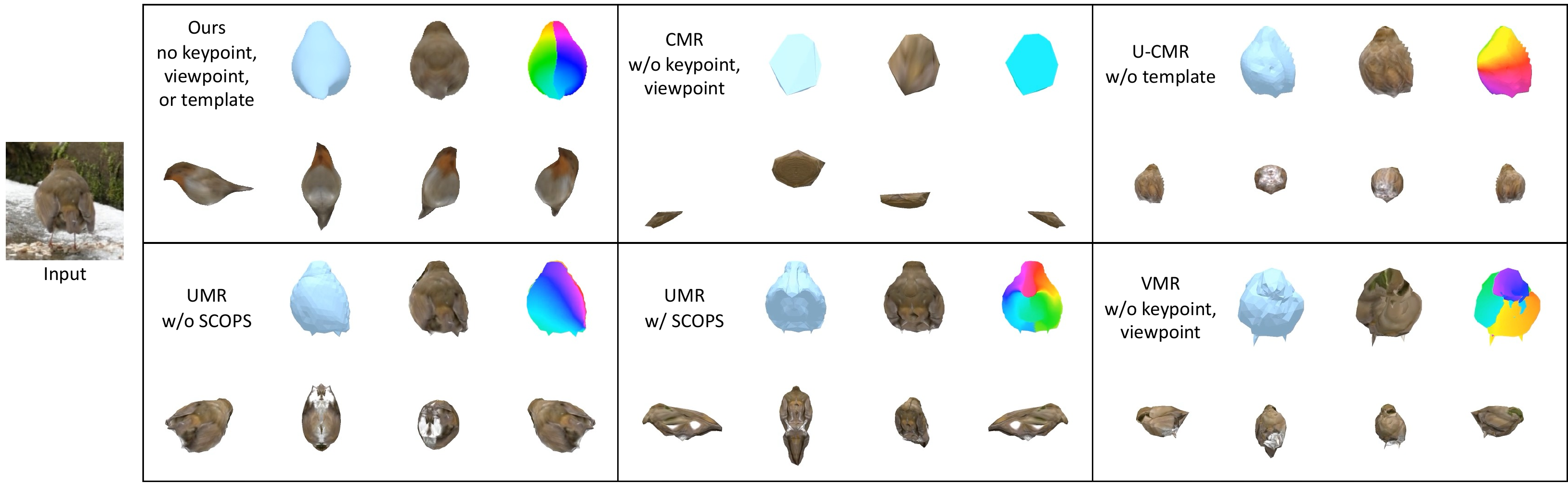}
    \includegraphics[width=\linewidth]{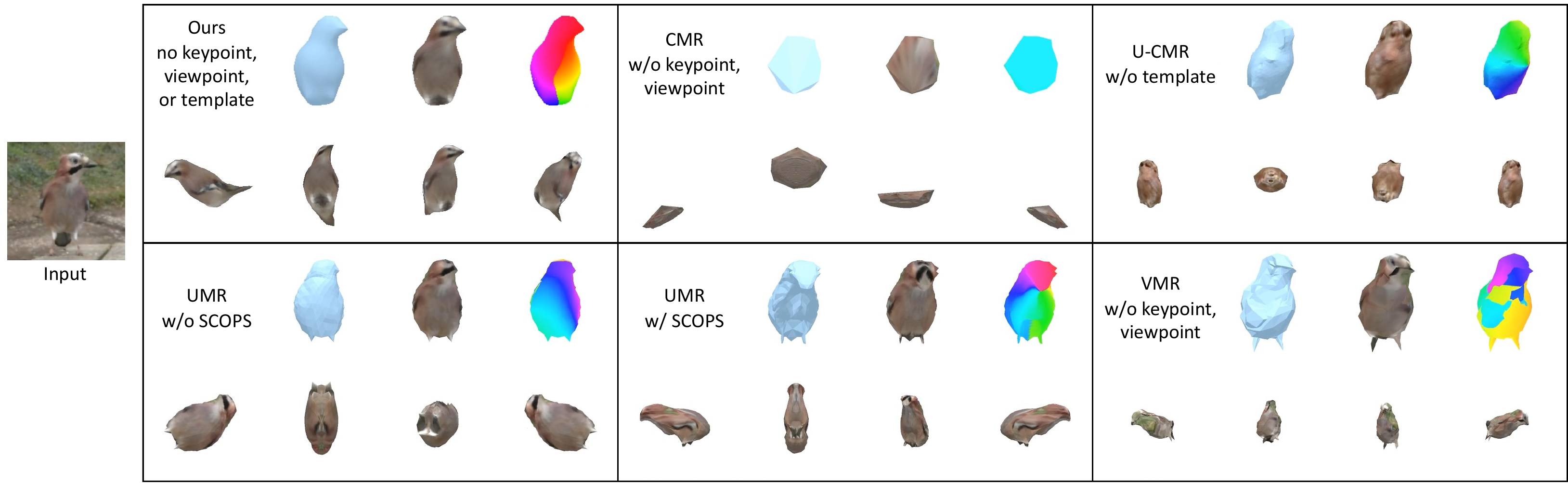}
    \includegraphics[width=\linewidth]{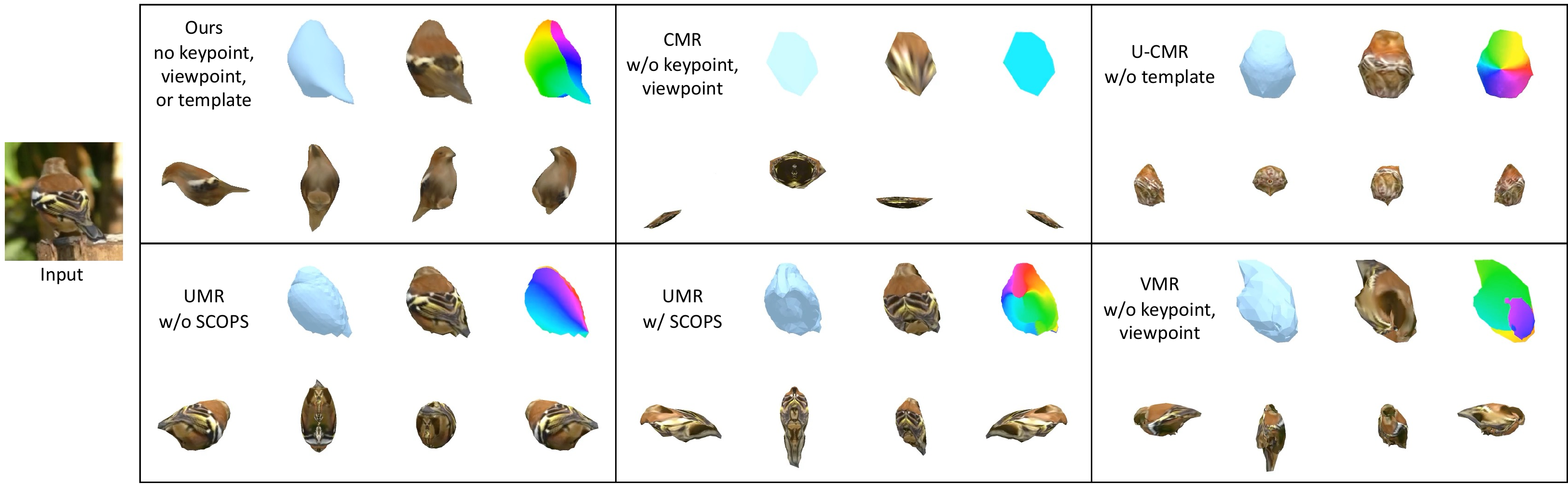}
    \includegraphics[width=\linewidth]{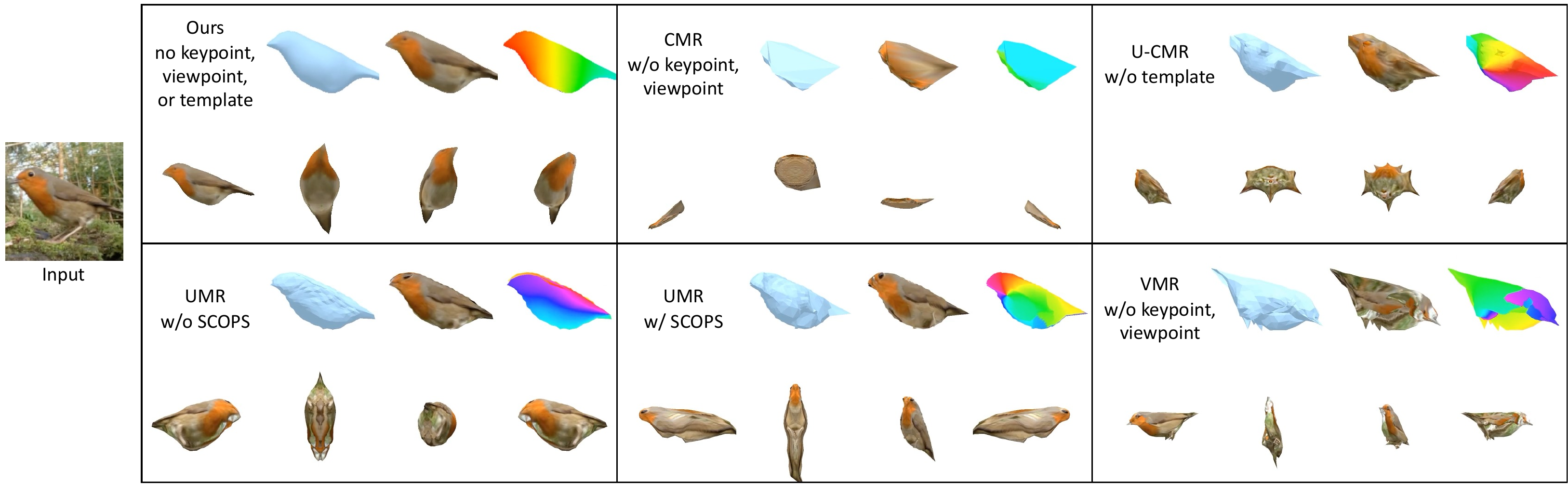}
    \caption{\textbf{Comparisons against Existing Methods.} Without additional geometric supervision, SOTA methods fail to learn correct poses and hence produces poor shape reconstructions, whereas our model learns more plausible 3D shapes and accurate poses.}
    \label{fig:supmat_compare}
\end{figure}

\paragraph{Comparison of Learned Rigid Pose Distributions.}
Estimating the viewpoint/object pose is a key factor of learning consistent 3D shapes.
We plot the elevation and azimuth of the rigid poses (viewpoint) predicted by various methods (CMR~\cite{kanazawa18learning}, U-CMR~\cite{goel20shape}, UMR~\cite{li20self-supervised} and VMR~\cite{li20online}) and compare them in \cref{fig:supmat_pose_compare}.
As shown in the plots, without additional geometric supervision (keypoint, viewpoint or template shape), existing methods are not able to learn correct poses and the predicted poses tend to collapse to only half of the azimuth, resulting in poor shape reconstructions that are not consistent across different inputs, as shown in \cref{fig:supmat_compare} as well as in the main paper.
The only exception is U-CMR~\cite{goel20shape} which explicitly uses an extensive viewpoint search (camera multiplex) to encourage diverse viewpoint predictions.
However, without a good shape template, the resulting viewpoints are also not correct leading to poor shapes.

\paragraph{Additional Qualitative Comparisons.}

\begin{wrapfigure}{r}{0.2\columnwidth}
\includegraphics[width=0.2\columnwidth]{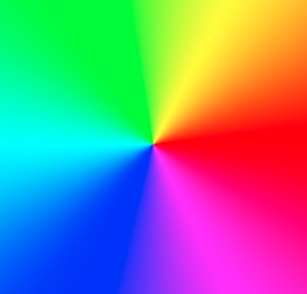}
\caption{Synthetic texture used for visualization in~\cref{fig:supmat_compare}.}
\label{fig:synth}
\end{wrapfigure}%
\cref{fig:supmat_compare} provides a few more examples comparing the reconstruction results of our model and several state-of-the-art methods.
We also show shape reconstructions rendered with a static synthetic texture, shown in~\cref{fig:synth}, to clearly illustrate the orientation of the predicted shape.
Our model learns more accurate 3D shapes, despite not requiring explicit geometric supervision from keypoints, viewpiont or template shapes.
More comparisons on entire video sequences are provided in the supplementary video.

\cref{fig:supmat_lasr_compare} shows more qualitative comparisons to LASR~\cite{yang21lasr:} on horses.

\subsubsection{PCA Analysis on Learned Shape Space}
\begin{figure}[t]
    \centering
    \includegraphics[trim={0 0 15px 0}, clip, width=\linewidth]{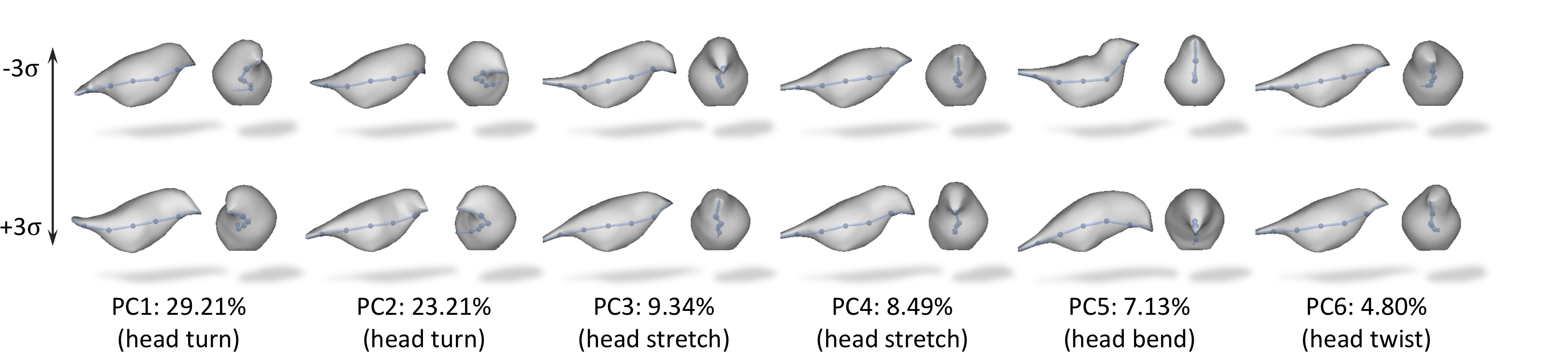}
    \caption{\textbf{PCA on Learned Articulations.} Our model learns meaningful articulations with the bone-based skinning model, including typical head and tail movements of birds.}
    \label{fig:art}
\end{figure}

We analyze the learned articulated shape space across the dataset using Principal Component Analysis (PCA).
\Cref{fig:art} visualizes the first $6$ principal components.
Each principal component corresponds to a typical bird movement, which means that the model learns meaningful articulations that are reflected in the skeleton and not in the shape deformation component.

\subsubsection{Additional Reconstruction Results}
\begin{figure}[t!]
    \centering
    \includegraphics[trim={0 0 15px 0}, clip, width=\linewidth]{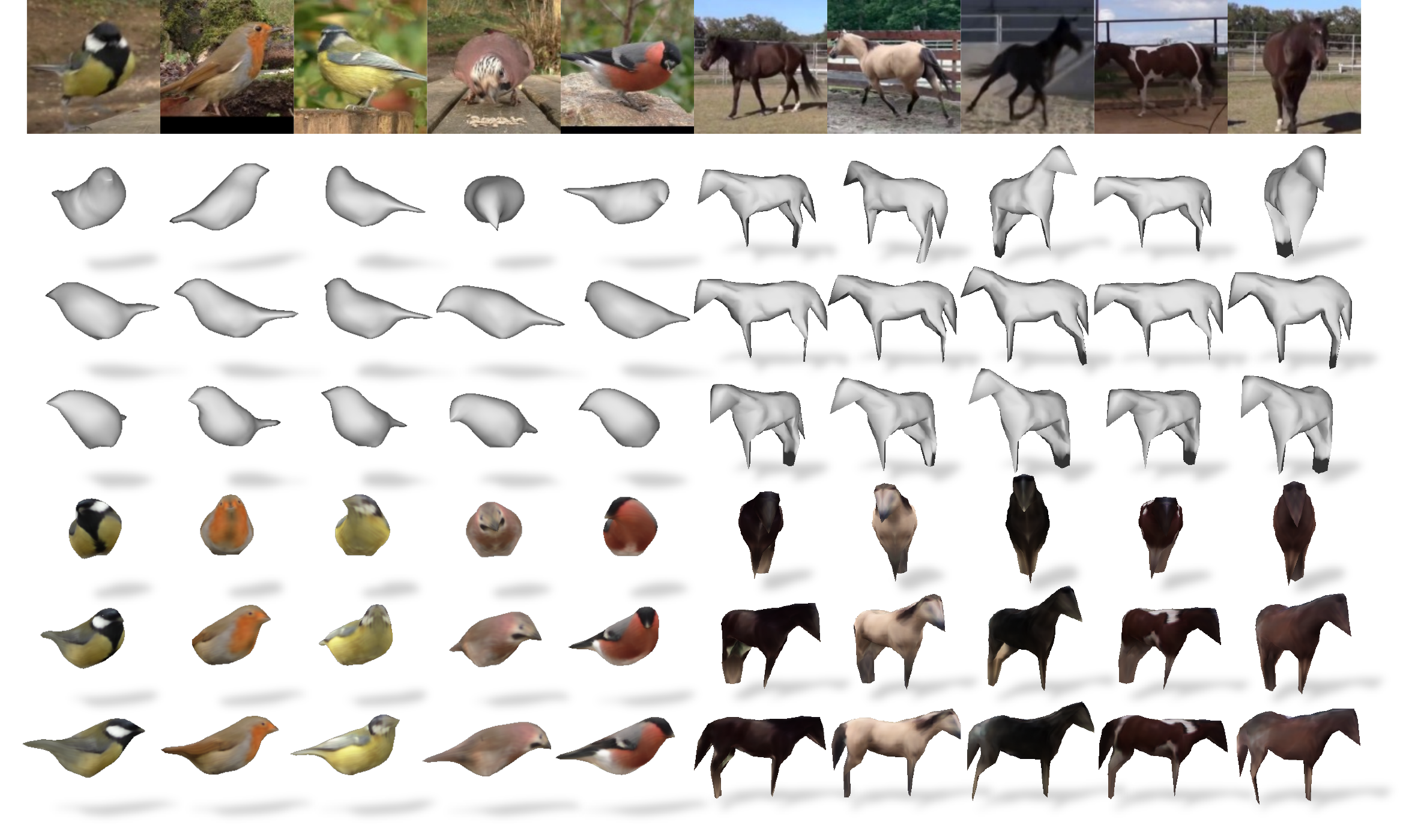}
    \caption{\textbf{Additional Reconstruction Results.} We show the reconstructed objects from various viewpoints.}
    \label{fig:supmat_recon}
\end{figure}

\Cref{fig:supmat_recon} shows more bird reconstructions from various viewpoints. The model is robust against various input images, including frontal views and blurry images.

\subsubsection{Texture Swapping and Animation}
\begin{figure}[t!]
    \centering
    \includegraphics[trim={0 0 15px 0}, clip, width=\linewidth]{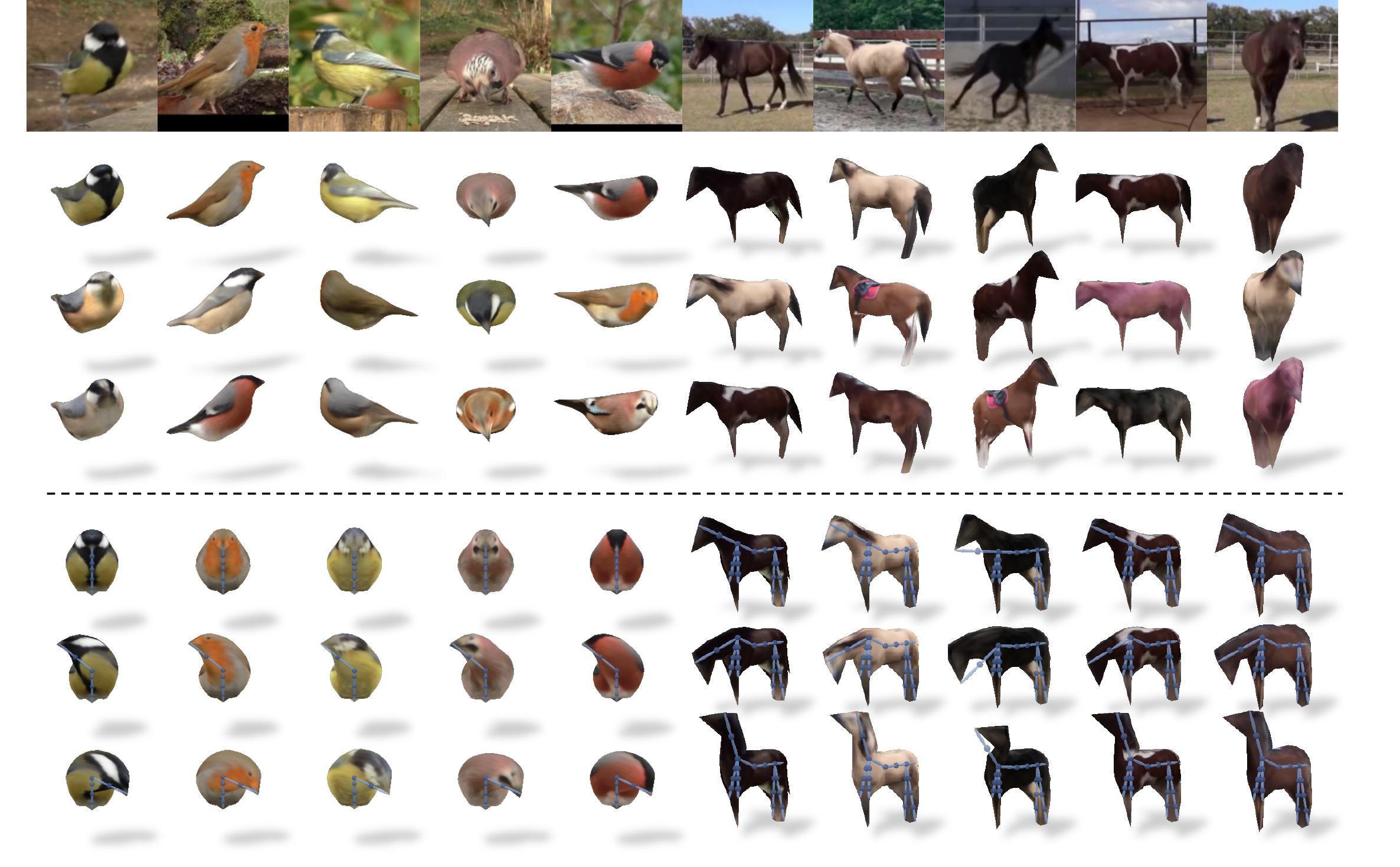}
    \caption{\textbf{Texture Swapping and Animation.} Top: since our model learns a canonical representation for all objects, we can easily swap the texture across different instances. Bottom: we can also easily animate the 3D birds using our learned articulation model.}
    \label{fig:swap_tex}
\end{figure}

Our model reconstructs the birds in the canonical pose, where the shape and texture of different birds are aligned in the canonical representation.
This allows us to easily edit the texture, for example swapping the texture with another bird, as shown in \cref{fig:swap_tex}.

Moreover, with the learned articulation model, we can also easily animate the reconstructed birds in 3D, by rotating the bones, also illustrated in \cref{fig:swap_tex}.

\subsection{Datasets}
We collected videos containing birds and horses from YouTube and automatically pre-processed them as follows.
We use PointRend model~\cite{kirillov2019pointrend} to obtain detection and segmentation of the object instances.
Each video is then split into short clips containing a single object while we also remove frames that would contain humans after final cropping.
We also filter out static frames from the clips using optical flow computed by off-the-shelf RAFT model~\cite{teed20raft:}.
The optical flow is then recomputed one more time to account for the removed frames.
Finally, we crop the frames, segmentation masks and optical flow around the detected object bounding boxes and resize them to $128 \times 128$ for training.

\paragraph{Bird Videos.}
We downloaded a single four-hour long video of birds with static camera from YouTube\footnote{\url{https://youtu.be/xbs7FT7dXYc}}.

\paragraph{Horse Videos}
We searched YouTube for terms such as \emph{horse training}, \emph{horse training in pen}, \emph{leading horse} and selected 12 videos
\footnote{\url{https://youtu.be/YM_jw4g0urc}, \url{https://youtu.be/0MSifR_4BAw}, \url{https://youtu.be/qincEZod6mQ}, \url{https://youtu.be/nyBMMoeg4CU}, \url{https://youtu.be/_BN-0e1r89E}, \url{https://youtu.be/Bb8nquNkruY}, \url{https://youtu.be/wq4vryHoe-0}, \url{https://youtu.be/asE5y2qO5dw}, \url{https://youtu.be/4ZuKilrSgXI}, \url{https://youtu.be/dJtyHyt6bOk}, \url{https://youtu.be/0YUAArukMFc}, \url{https://youtu.be/U61UqV0x4QQ}}
where a horse is trained without a mounted rider.
As the content in these videos can be varied, we manually roughly split the videos into clips where the horse is the main focus before applying the automatic pre-processing.

\subsubsection{3D Toy Bird Dataset}
\begin{figure}[t!]
    \includegraphics[width=1\linewidth]{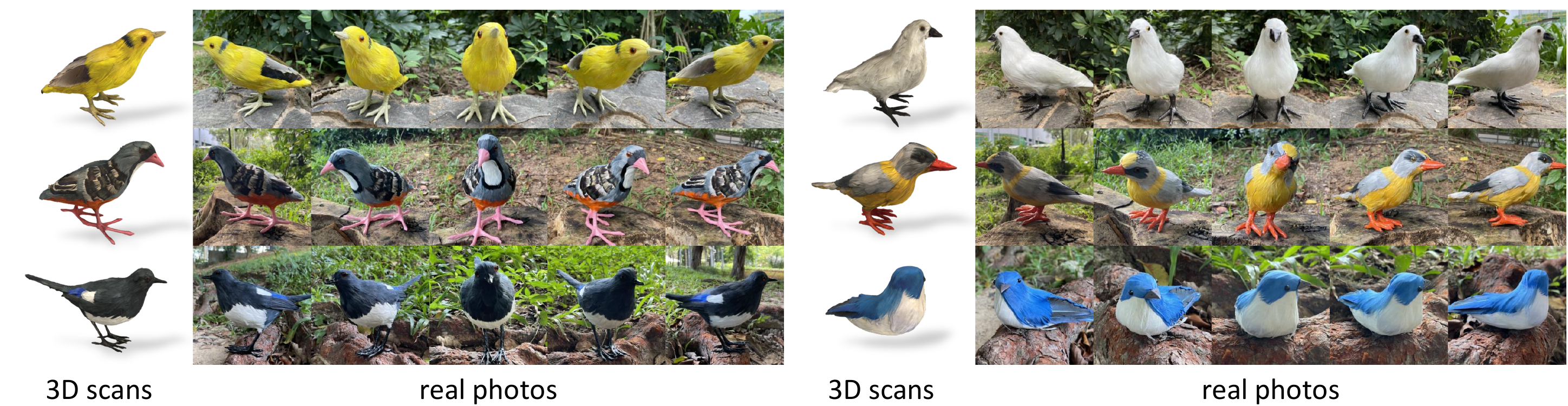}
    \caption{\textbf{More examples of the 3D Toy Bird Dataset.}
    }
    \label{fig:supmat_toy_bird}
\end{figure}

\Cref{fig:supmat_toy_bird} shows renderings of six 3D models in our 3D toy bird dataset and the corresponding photographs of the toys ``in the wild''.
The visual appearance of the birds in the photos is close to the real one is CUB200 or our video dataset. The 3D reconstructions are of high quality and allow for precise quantitative evaluation.
For future benchmarking and comparisons, we will release the dataset together with the paper.

\subsection{Broader Impact}
Our work focuses on 3D reconstruction of deformable objects from monocular videos.
We expect this work to be most useful for object categories that do not have sophisticated 3D ground-truth annotations and 3D shape models.
This is mainly the case for animals, as shown for birds and horses here, and thus the work can potentially be of use in behavioral research of animals in the wild.
Our 3D Toy Bird dataset does not contain any humans, only toy birds in nature background and was captured by the authors. Thus, the copyright of the dataset is with the authors and it does not violate the personal privacy of individuals.
Overall, we expect this work to impact mostly the research community with very little impact on society in the short term.

In the long term, we expect the task of obtaining 3D models and algorithms that can lift objects from images into 3D to become more and more important; for example in XR and VFX applications. Our method shows that obtaining these models can be achieved without external geometric supervision, which is often difficult to collect for arbitrary object categories.

\begin{table}
\caption{Architecture of the shape $f_S$ and pose network $f_P$. The network follows a convolutional encoder structure. $n$ is the number of parameters predicted by each network.}
\label{tab:arch}
\centering
\begin{tabular}{lc}
\toprule
 Encoder & Output size \\ \midrule
 Conv(3, 64, 4, 2, 1) + GN(16) + LReLU(0.2) & 64 $\times$ 64\\
 Conv(64, 128, 4, 2, 1) + GN(32) + LReLU(0.2) & 32 $\times$ 32\\
 Conv(128, 256, 4, 2, 1) + GN(64) + LReLU(0.2) & 16 $\times$ 16\\
 Conv(256, 512, 4, 2, 1) + GN(128) + LReLU(0.2) & 8 $\times$ 8\\
 Conv(512, 512, 4, 2, 1) + LReLU(0.2) & 4 $\times$ 4\\
 Conv(512, 256, 4, 1, 0) + ReLU & 1 $\times$ 1\\ 
 Conv(256, $n$, 1, 1, 0) $\rightarrow$ output & 1 $\times$ 1\\ 
\bottomrule
\end{tabular}
\end{table}

\begin{table}
\caption{Architecture of the texture network $f_T$. The network follows an encoder-decoder structure.}
\label{tab:arch_tex}
\centering
\begin{tabular}{lc}
\toprule
 Encoder & Output size \\ \midrule
 Conv(3, 64, 4, 2, 1) + GN(16) + LReLU(0.2) & 64 $\times$ 64\\
 Conv(64, 128, 4, 2, 1) + GN(32) + LReLU(0.2) & 32 $\times$ 32\\
 Conv(128, 256, 4, 2, 1) + GN(64) + LReLU(0.2) & 16 $\times$ 16\\
 Conv(256, 512, 4, 2, 1) + GN(128) + LReLU(0.2) & 8 $\times$ 8\\
 Conv(512, 512, 4, 2, 1) + LReLU(0.2) & 4 $\times$ 4\\
 Conv(512, 256, 4, 1, 0) + ReLU & 1 $\times$ 1\\ \midrule \midrule
 Decoder & Output size \\ \midrule
 Deconv(256, 512, 4, 1, 0) + ReLU & 4 $\times$ 4\\
 Upsample(2) + Conv(512, 512, 3, 1, 1) + GN(128) + ReLU & 8 $\times$ 8\\
 Upsample(2) + Conv(512, 256, 3, 1, 1) + GN(64) + ReLU & 16 $\times$ 16\\
 Upsample(2) + Conv(256, 128, 3, 1, 1) + GN(32) + ReLU & 32 $\times$ 32\\
 Upsample(2) + Conv(128, 64, 3, 1, 1) + GN(16) + ReLU & 64 $\times$ 64\\
 Upsample(2) + Conv(64, 64, 3, 1, 1) + GN(16) + ReLU & 128 $\times$ 128\\
 Conv(64, 3, 5, 1, 2) + Sigmoid $\rightarrow$ output $T$ & 128 $\times$ 128\\
\bottomrule
\end{tabular}
\end{table}

\begin{table}
\begin{minipage}[t]{.47\linewidth}
\caption{Training details and hyper-parameters.}
\label{tab:params}
\centering
\begin{tabular}{lc}
\toprule
 Parameter & Value/Range \\ \midrule
 Optimizer & Adam \\
 Learning rate ($f_V$, $f_\xi$, $f_T$) & $1\times 10^{-4}$ \\
 Learning rate ($\Delta V_\text{tmpl}$) & $1\times 10^{-2}$ \\
 Number of epochs & $16$ \\
 Number of sequences per batch & $4$ \\
 Number of frames per sequence & $8$ \\
 Loss weight $\lambda_{\text{im}}$ & $1$ \\
 Loss weight $\lambda_{\text{mask}}$ & $2$ \\
 Loss weight $\lambda_{\text{ARAP}}$ & $50$ \\
 Loss weight $\lambda_{\text{Lap}}$ & $1$ \\
 Loss weight $\lambda_{\text{nrm}}$ & $1$ \\
 Loss weight $\lambda_{\text{flow}}$ & $100$ \\
 Loss weight $\lambda_{\text{sym}}$ & $0.05$ \\
 Loss weight $\lambda_{\text{pose}}$ & $0.1$ \\
 \midrule
 Input image size & $128 \times 128$ \\
 Texture image size & $128 \times 128$ \\
 Field of view (FOV) & $25^\circ$ \\
 Camera location & $(0, 0, 10)$ \\
 Initial mesh center & $(0, 0, 0)$ \\
 Number of bones (bird) & $6$ \\
 Number of bones (horse) & $6+4\times 4$ \\
\bottomrule
\end{tabular}
\end{minipage}
\hfill
\begin{minipage}[t]{.47\linewidth}
    \centering
    \captionof{figure}{\textbf{Texture Mapping.} The texture is mapped from a circle in the texture to both, left and right, sides of the initial sphere.}
    \vspace{0.05in}
    \includegraphics[width=0.95\linewidth]{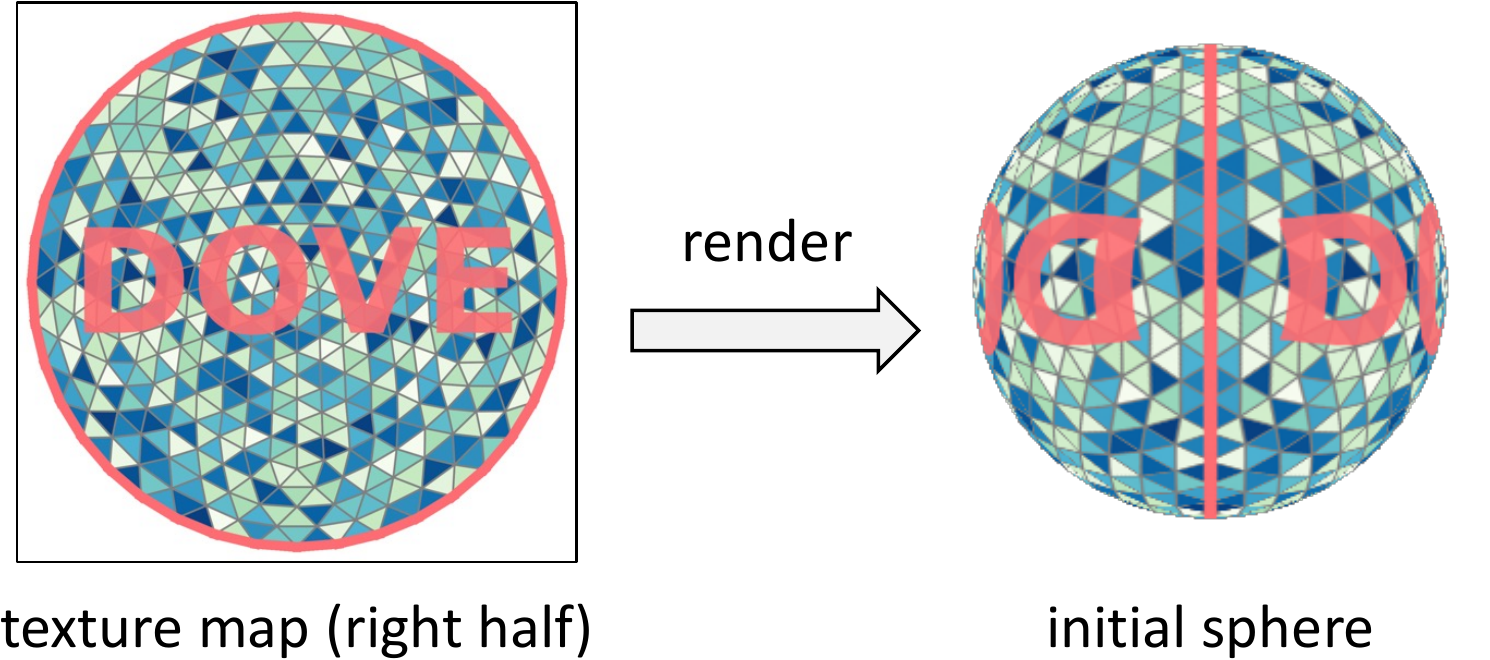}
    \label{fig:tex_map}
\end{minipage}
\end{table}

\subsection{Mathematical Details}

In this section we expand the underlying math of shape, pose and symmetries to provide a detailed mathematical foundation of the underlying principles.

\subsubsection{Shape and Pose}
Recall that we represent the shape of each object instance using a hierarchical shape model.
The model first predicts the shape of each instance $V$ at rest pose as:
\begin{equation}
    V = V_\text{base} + \Delta V_\text{tmpl} + \Delta V,
\end{equation}
where $V_\text{base}$ is the initial fixed shape parametrized by a symmetric ico-sphere mesh with $642$ vertices and $1,280$ faces.
$\Delta V_\text{tmpl}$ is a set of trainable parameters initilized as zeros and directly optimized during training, such that $V_\text{tmpl} = V_\text{base} + \Delta V_\text{tmpl}$ represents the category-specific template shape shared across all instances of the category.
$\Delta V$ is instance-specific deformation predicted from each input image by the shape network $f_V$.

This rest pose shape $V$ is further transformed into the actual mesh observed in the image by a posing (or skinning) function $g(V, \xi)$, described in~\cref{sec:skinning}, where the pose parameter $\xi$ consists of the rigid pose $\xi_1 \in SE(3)$ and a set of rotations of the bones $\xi_b \in SO(3), b=2,...,B$.

The rigid pose is predicted by the pose network $f_\xi$ as a 3D rotation parametrized by a forward vector and 3D translations in $xyz$ axes.
Specifically, we predict a forward vector and define $(0, 1, 0)^T$ as the up direction and obtain a basis for the rotation matrix from these vectors by two consecutive cross products.
This effectively disables the in-plane rotation as the objects tend to stay mostly upright.
The translations are capped at a range roughly corresponding to $0.4$ of the image size after projection.

The bone rotations are predicted by the shape network $f_V$ simply as a set of Euler angles.

\subsubsection{Skinning Equation}
\label{sec:skinning}
Each bone $b$ is rigidly attached to a parent bone $\pi(b)$ forming a kinematic tree.
Vector $\mathbf{J}_b$ is the location of the joint between $b$ and $\pi(b)$ expressed relative to the parent.
If $V_{ib}$ are the coordinates of mesh vertex $i$ expressed relative to bone $b$, we can find the location in world space as:
$$
 V_{i0} = G_b(\xi) V_{ib},
 ~~
 G_b = G_{\pi(b)} \circ g_b,
 ~~
 g_b(\xi) = \begin{bmatrix}
    R_{\xi_b} & \mathbf{J}_b \\ 0 & 1 \\
\end{bmatrix}.
$$
Given an initial mesh $V$ at rest, then posed version is found by first expressing the vertices relative to each bone $b$ in the rest configuration $\xi_0$ using transformations $G_b(\xi_0)^{-1}$ and then posing the vertex by applying transformation $G_b(\xi)$.
This is weighed by the strength of association of each vertex to a bone, resulting in the \emph{skinning equation}~\cite{loper15smpl:}:
\begin{equation}\label{e:skinning}
V_i(\xi) =\left( \sum_{b=1}^B w_{ib} G_b(\xi) G_b(\xi_0)^{-1} \right) V_i.
\end{equation}

\paragraph{Skinning Weights.}
As described in the main paper, we estimate the bone structure using the two most extreme points of the mesh at its rest pose.
The one with positive $z$ coordinate is selected as the head end and the other one the tail end, ensuring consistent orientation of the bone structure.
The rotation of individual bones is represented using Euler angles about the $xyz$-axes in its local coordinate frame, where the center of rotation is specified by the joint location.

Recall Eq.~(2) of the main paper where we use skinning weights $w_{bi}$ that associate each mesh vertex with the bones.
We softly assign each mesh vertex with the bones based on its distances to the bones.
The weights are defined as the inverse of the distance between a vertex $[V_\text{ins}]_i$ and a bone $b$ at their rest pose.
We normalize this distance for each vertex over all the bones with softmax function with temperature:
\begin{equation}\label{e:skinning_w}
  w_{bi} = \frac{e^{d_{bi}/T}}{\sum_{k=1}^{B}{e^{d_{ki}/T}}},
  ~~~~
  d_{bi} = 1 / (\min_{r \in [0,1]} \|[V_\text{ins}]_i - r\mathbf{s}_{b1} - (1-r) \mathbf{s}_{b2}\|^2_2 + \epsilon),
\end{equation}
where $T$ is the temperature parameter, $(\mathbf{s}_{b1}, \mathbf{s}_{b2})$ is the line segment defining the bone $b$ at its rest pose and $\epsilon$ is a small number to avoid division by zero.

\paragraph{Bone Structure Initialization.}
We estimate the bone structure using a simple heuristics.
Given the shape corresponding to the rest pose, we define a fixed number of bones inside the mesh forming a `spine' which we set to lie on two line segments going from the two most extreme point of the mesh to the center.
We then divide each line segment into equally-sized parts that define the origin and the length of each bone.

For quadruped animals, we also define bones forming four legs.
We divide the space into quadrants along the $x$ and $z$ axis and then we found the lowest point on the mesh along the $y$ axis in each quadrant.
These lowest points are supposed to correspond to feet.
We define a line segment between each of the lowest points, and the closest joint on the spine.
As done with the spine, we then divide each line segment into equally-sized parts forming the individual bones.

\subsubsection{Symmetry}
We first define the action of the mirror mapping $m$ on several objects.
First, $m$ acts on 3D points as the matrix that flips the $x$ axis:
\begin{equation}\label{e:m-basic}
    m =
    \begin{bmatrix}
        -1 &&\\&1&\\&&1
    \end{bmatrix}.
\end{equation}
Note that $m^{-1}=m$.

Next, let $V=\{V_i \in \mathbb{R}^3\}_{i=1,\dots,|V|}$ be the vertices of a mesh.
The mesh is posed by the transformation $g \in SE(3)$, which we also write as $g(\cdot, \xi)$ for a certain vector of pose parameters $\xi$.
The \emph{mirror-symmetric pose} of $g$ is the \emph{conjugate transformation} $mgm \in SE(3)$.
This also defines the meaning of the symbol $m\xi$ as the parameters of the conjugate:
$$
     g(\cdot, m \xi) \equiv m g(\cdot, \xi) m.
$$
Thus, given a point $X \in \mathbb{R}^3$, this definition satisfies the equation
$
m g X = (m g m) m X
$
or:
$$
  m g(X, \xi) = g(m X, m \xi).
$$
This means that posing the point $X$ and mirroring the result is the same as applying the mirror pose to the mirror point $m X$.

Now let $T_i$ be the color of vertex $V_i$.
This determines the color $I(u) = T_i$ of pixel $u \in \mathbb{R}^2 \times \{1\}$ where
$$
u \propto  g V_i
$$ 
is the perspective projection of the posed vertex $V_i$ onto the image plane.
If we mirror pixel $u$, we obtain:
$$
m u \propto m g V_i = (m g m) (m V_i).
$$
If we define the \emph{mirror image} $(mI)(u) = I(mu)$ as the flipped copy of image $I$, and if $I$ is the image of object $(V,g,T)$, then $mI$ is the image of the mirrored object $(mV, mgm, T)$.

Next, we assume that the mesh is symmetric.
This means that every vertex $V_i$ has a symmetric counterpart $V_{m(i)}$, where $m$ acts as a permutation on the vertex indices:
$$
\forall i \in \{1,\dots,|V|\} :~~~  m V_i = V_{m(i)}.
$$
In fact this is not an arbitrary permutation: it consists of swaps, meaning that $m(m(i)) = i$, so that we have $m^{-1} = m$, just as before.

If is convenient to define the right action $m$ on the vertex collection $V$ as applying this permutation.
Using matrix notation, if we interpret $V$ as a $3\times |V|$ matrix, this means that in the expression $mV$ the operator $m$ acts as the axis-flipping operator~\cref{e:m-basic}, and in the expression $Vm$ it acts a $|V| \times |V|$ permutation matrix.
With this convention, the mesh is symmetric if, and only if,
$$
m V = Vm.
$$

\paragraph{Mirror Equivariance.}

To summarise, if $I$ is the image of $(V,g,T)$, them $mI$ is the image of $(mV, mgm, T)$ in general, and of $(Vm, mgm, T)$ if the mesh is symmetric.
Because the order of the mesh vertices is irrelevant for rendering an image, $mI$ is also the image of object $(V mm^{-1}, mgm, Tm^{-1})= (V, mgm, Tm)$.
In other words, we obtain the mirror image $mI$ by viewing the same symmetric shape $V$ under a mirror pose and texture.
If the texture is also symmetric (i.e, $Tm =T$), then we only need to mirror the pose.

We conclude that, if the network makes the prediction $(V,g,T) = f(I)$ for image $I$, then it mus make the prediction $V, mgm, Tm) = f(mI)$ for the mirror image.

\paragraph{Front-to-Back Ambiguity.}

Next, instead of a perspective camera, consider an orthographic projection
$$
 \Pi = \begin{bmatrix} 1 &&\\ &1&\\ &&0\end{bmatrix}
$$
so that vertex $V_i$ projects to point $u = \Pi V_i$ in the image.
The same consideration as before apply without changes.
Additionally, consider the rotation matrix
$$
r =
\begin{bmatrix}
    -1 &&\\&1&\\&&-1
\end{bmatrix}.
$$
which consists of a 180 degree rotation around the $y$ axis.
Given a symmetric mesh $mV = Vm$, consider objects $(V, g, T)$ and $(Vm, rmgm, T)$.
If we pose the second, we see that:
$$
(rmgm) (Vm) =  (rm)(gV)
=
\begin{bmatrix}
    1 &&\\&1&\\&&-1
\end{bmatrix} (gV).
$$
This means that the posed points are exactly the same, except for the fact that the $z$ (not $x$!) axis is inverted.
If we image them using the orographic projection $\Pi$, this axis is removed, so the image points are exactly the same in the two cases.
Naturally, because the depth changes, the images are different; however, the masks are identical, causing the ambiguity.

Ton conclude, this means that the network $(V,g,T)=f(I)$ is likely to be confused between pose $g$ and pose $rmgm$.
Hence, we define $q\xi$ so that:
$$
  rmg(\dot, \xi)m = g(\cdot, q\xi).
$$

\paragraph{Extension to Articulated Pose.}

Fortunately, the discussion above generalizes to articulated poses.
Posing uses in fact a chain of $SE(3)$ transformations of the type:
$$
 G_b(\xi) G_b(\xi)^{-1},
 ~~~~ \text{where}~~G_b(\xi) = G_{\pi(b)}(\xi) \cdot g_b(\xi_b).
$$
The conjugate $mG_b(\xi) G_b(\xi)^{-1}m$ is found by interjecting factors $mm =1$ in the chain:
$$
m(G_{\pi(b)}(\xi) \cdot g_b(\xi_b))m
=
(mG_{\pi(b)}(\xi)m)\cdot(mg_b(\xi_b)m)
$$
Expanding the last term we get that:
$$
(mg_b(\xi)m)(X) = m \begin{bmatrix}
    R_{\xi_b} & \mathbf{J}_b \\ 0 & 1 \\
\end{bmatrix}(mX)
=
m R_{\xi_b} m mX + m  \mathbf{J}_b.
$$
We assume that the bone structure is also symmetric, so that we can define for each bone $b$ a symmetric counter-part $m(b)$ (another permutation) such that:
$$
  m \mathbf{J}_b = \mathbf{J}_{m(b)}.
$$
With this, we have that:
$$
(mg_b(\xi)m)(X) = g_{m(b)}(m\xi_{m(b)}) (mX)
= g_{m(b)}((m\xi m)_{b}) (mX).
$$
Likewise, we assume the symmetry of the skinning weights and of the base vertices:
$$
w_{ib}= w_{m(i), m(b)},
~~~~
m V_i = V_{m(i)}.
$$
With this:
\begin{align*}\label{e:sym-skinning}
m V_i(\xi) &= m \left(\sum_{b=1}^B w_{ib} G_b(\xi) G_b(\xi_0)^{-1} \right) m mV_i\\
           &= \left(\sum_{b=1}^B w_{ib} (m G_b(\xi) m)(m G_b(\xi_0) m)^{-1}\right) V_{m(i)} \\
           &= \left(\sum_{b=1}^B w_{m(i),m(b)} G_{m(b)}(m\xi m) G_{m(b)}(m\xi_0m)^{-1} \right) V_{m(i)} \\
           &= \left(\sum_{b=1}^B w_{m(i)b} G_{b}(m\xi m) G_{b}(m\xi_0m)^{-1} \right) V_{m(i)} \\
           &= V_{m(i)} (m\xi m).
\end{align*}
The equation $m V_i(\xi) = V_{m(i)} (m\xi m)$ is a generalization of the equation
$
m g V_i = (mgm) V_{m(i)}
$
found before.

\subsection{Implementation Details}

\subsubsection{Texture}
We assume the texture of the objects is symmetric, and predict a texture map $T$ for each object.
Each vertex is associated with a fixed uv texture coordinate defined on the initial sphere, such that the texture is mapped from a circle in the texture map to both, left and right, sides of the sphere, as illustrated in~\cref{fig:tex_map}.
Mathematically, this is computed as:
\begin{equation}
    u = \frac{2\arccos{|x|}}{\pi} \frac{z}{\sqrt{z^2+y^2}},
    ~~~~
    v = \frac{2\arccos{|x|}}{\pi} \frac{y}{\sqrt{z^2+y^2}},
\end{equation}
where $x$, $y$ and $z$ are the 3D coordinates of the vertices of a unit sphere.

\subsubsection{Renderer Details}
We use Pytorch3D differentiable renderer~\cite{ravi20accelerating}.
We use a perspective camera with a FOV of 25$^\circ$, that is assumed to be stationary at $(0, 0, 10)$ looking at $(0, 0, 0)$.
The initial sphere is a unit sphere centered at $(0, 0, 0)$.
If the background image can be easily extracted from the video, \eg videos with a static camera, we overlay the rendered mesh onto the extracted background, and compute the photometric loss described below on the entire image.
Otherwise, we compute it only on the foreground pixels.

\subsubsection{Geometric Smoothing Terms}

In order to prevent the instance-specific shapes $V$ to deviate too much from the category prior $V_\text{tmpl}$, we use the As-Rigid-As-Possible (ARAP) regularizer~\cite{sorkine2007rigid}:
\begin{equation}
    L_{\text{ARAP}}
    = \sum_{i=1}^{|V|} 
    \min_{R\in SO(3)}
    \sum_{j \in \mathcal{N}_{i}} 
    \| 
    (V_{i} - V_{j})
    - R 
    (V_{\text{tmpl}, i} - V_{\text{tmpl}, j}) 
    \|^2,
\end{equation}
where $\mathcal{N}_k$ is the fan of vertices around vertex $k$.
ARAP thus encourages triangle fans to deform rigidly.
We further apply Laplacian and normal smoothing losses to the posed mesh $V_\xi$:
\begin{align}
L_{\text{Lap}} &= 
\sum_{i=1}^{|V_\xi|} 
\Big\|
V_{\xi i} 
- 
\frac{1}{|\mathcal{N}_i|} 
\sum_{j \in \mathcal{N}_i}
V_{\xi j}
\Big\|^2,
~~~
L_{\text{nrm}} =
\sum_{f\in\mathcal{F}}
\sum_{f'\in\mathcal{N}_f}
\left(
1-
\frac
{\mathbf{n}_{if} \cdot \mathbf{n}_{if'}}
{\|\mathbf{n}_{if}\| \|\mathbf{n}_{if'}\|}
\right),
\end{align}
where $\mathbf{n}_{if}$ is the normal of face $f$ in $V_\xi$ and $\mathcal{N}_f$ are the faces adjacent to $f$ in the mesh.
We sum the three geometric losses in the shape regularizer
$
L_{\text{smooth}}(V_\text{tmpl}, V, V_\xi)
= 
\lambda_\text{ARAP} L_{\text{ARAP}} +
\lambda_\text{Lap} L_{\text{Lap}} +
\lambda_\text{nrm} L_{\text{nrm}}.
$

\subsubsection{Network Architectures}

All the network architectures are described in in~\cref{tab:arch,tab:arch_tex}.
Abbreviations of the components are defined as follows:

\begin{itemize}
	\item $\text{Conv}(c_{in}, c_{out}, k, s, p)$: 2D convolution with $c_{in}$ input channels, $c_{out}$ output channels, kernel size $k$, stride $s$ and padding $p$;
	\item $\text{Deconv}(c_{in}, c_{out}, k, s, p)$: 2D deconvolution with $c_{in}$ input channels, $c_{out}$ output channels, kernel size $k$, stride $s$ and padding $p$;
	\item $\text{Upsample}(s)$: 2D nearest-neighbor upsampling with a scale factor $s$;
	\item $\text{GN}(n)$: group normalization~\cite{wu2018group} with $n$ groups;
	\item $\text{LReLU}(p)$: leaky ReLU~\cite{maas2013rectifier} with a slope $p$;
\end{itemize}

\subsubsection{Training Details}

All hyper-parameters are specified in \cref{tab:params}.
The model is trained for $20$ epochs, which takes three days on one NVIDIA RTX-6000 GPU.
For each epoch, we iterate through the training set by densely sampling $4$ short sequences in a batch for each iteration.
Each sequence contains $8$ consecutive frames.

We train the model is three phases.
(1) In the first phase, the goal is to learn a category template shape as well as rough as rough rigid poses of the objects. Therefore, we train the basic model with no instance-specific deformation, no bone articulation and no texture loss for the first epochs.
(2) During the second phase,
once the initial shape and rigid pose is learned, we activate the pose rectification loss $L_\text{pose}$ rectify incorrect front/back poses, for one epoch.
(3) Finally, we train the full model with instance deformation, skinning model with the bones instantiated, as well as texture loss for another 13 epochs.
Furthermore, we gradually decay the weight of the Laplacian and normal smoothness regularizers by $0.7$ every epoch from $1$ to the $0.05$, to allow the model to learn the geometric details in a coarse-to-fine manner.

To better enforce temporal consistency, we randomly replace both the instance-specific rest-pose shape and texture with their sequence averages with a $50\%$ probability.

We use Adam optimizers with a learning rate of $0.0001$ for the networks and $0.01$ for the trainable category template shape parameters $\Delta V_\text{tmpl}$.
The learning rates are decayed by a factor of $0.7$ after each epoch starting after 8 epochs.

\paragraph{Technical Details for SOTA Baselines.}
We used the open-sourced implementations and pre-trained models of CMR~\cite{kanazawa18learning}, U-CMR~\cite{goel20shape}, UMR~\cite{li20self-supervised}. We used a code and models of VMR~\cite{li20online} provided by the authors upon request.
We finetuned their published models pre-trained on CUB dataset by training them on our bird video dataset.
In the case of UMR, we also trained a version from scratch using only our birds dataset with SCOPS model pre-trained on CUB dataset.

We also trained all the baselines from scratch on our birds dataset without the additional supervisions that they need to use by default.
We initialized CMR, U-CMR and VMR with a sphere base shape instead of their template shapes.


\begin{figure}[t!]
    \includegraphics[width=1\linewidth]{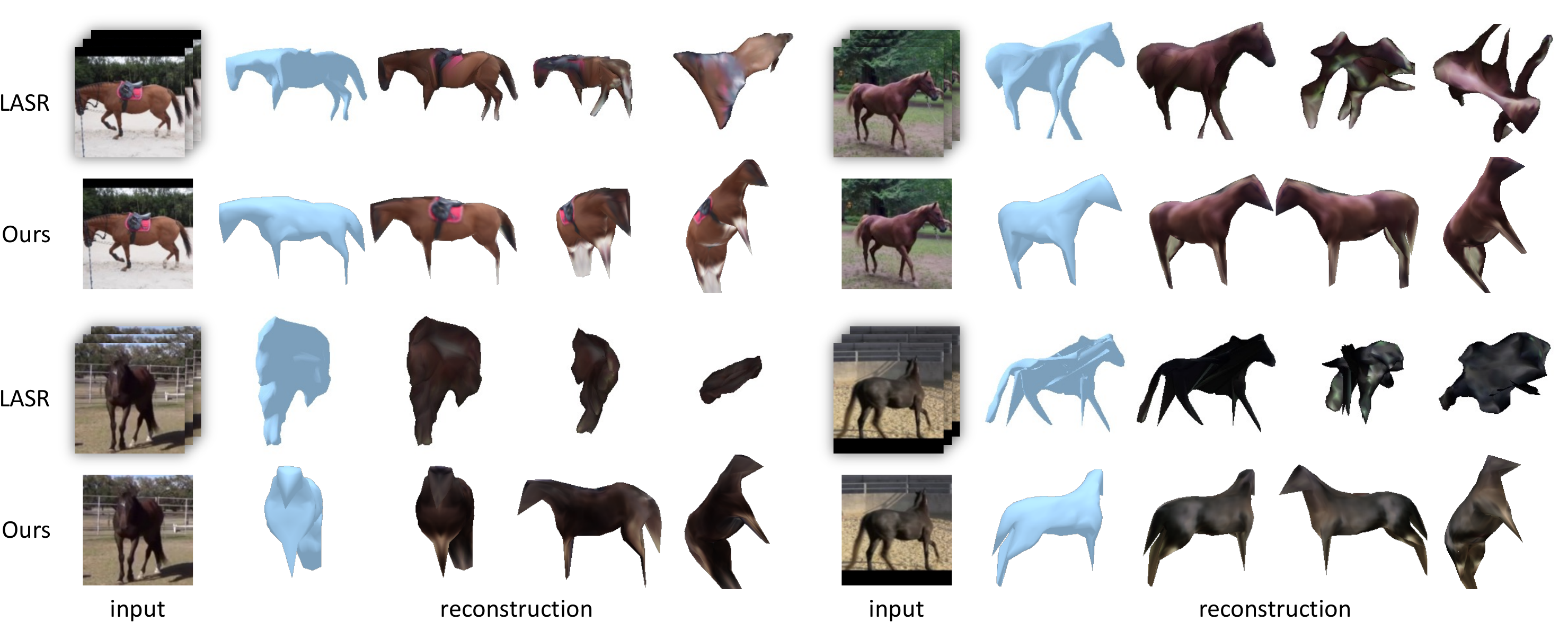}
        \caption{\textbf{Additional qualitative comparison with LASR~\cite{yang21lasr:}.} The shape produced by LASR is distorted and does not resemble the actual shape of a horse. Since our method trains on multiple sequences it learns consistent shapes of horses.}
    \label{fig:supmat_lasr_compare}
\end{figure}


\end{document}